\documentclass[journal]{IEEEtran}

\ifCLASSINFOpdf
   \usepackage[pdftex]{graphicx}
\else
   \usepackage[dvips]{graphicx}
\fi

\usepackage{epsfig}
\usepackage{amsmath}
\usepackage{amssymb}
\usepackage{subfigure}
\usepackage{booktabs}
\usepackage{tabularx}
\usepackage{nonfloat}
\usepackage{color}
\usepackage{graphicx}
\usepackage[numbers,sort&compress]{natbib}
\usepackage[normalem]{ulem}
\usepackage{multirow}
\usepackage{rotating}
\usepackage{multirow}
\usepackage[ruled,linesnumbered]{algorithm2e}
\usepackage{bbm}
\usepackage{bbding}
\usepackage{ragged2e}
\usepackage{algorithmic}
\usepackage{balance}
\usepackage{url}
\newcommand{\PreserveBackslash}[1]{\let\temp=\\#1\let\\=\temp}
\newcolumntype{C}[1]{>{\PreserveBackslash\centering}p{#1}}
\newcolumntype{R}[1]{>{\PreserveBackslash\raggedleft}p{#1}}
\newcolumntype{L}[1]{>{\PreserveBackslash\raggedright}p{#1}}

\newcommand{\figref}[1]{Figure~\ref{fig:#1}}

\newlength\secmargin
\newlength\subsecmargin
\newlength\paramargin
\newlength\figmargin
\newlength\eqmargin
\ExplSyntaxOn
\newcommand\latinabbrev[1]{
	\peek_meaning:NTF . {
		#1\@}%
	{ \peek_catcode:NTF a {
			#1.\@ }%
		{#1.\@}}}
\ExplSyntaxOff
\def\eg{\latinabbrev{e.g}}
\def\etal{\latinabbrev{et al}}

\def\ie{\latinabbrev{i.e}}

\begin{document}


\renewcommand{\baselinestretch}{0.998}

\newenvironment{myitemize2}[1][]{
\begin{list}{$\bullet$}
    {
     \setlength{\leftmargin}{5mm} 
     \setlength{\parsep}{0.5mm} 
     \setlength{\topsep}{0mm} 
     \setlength{\itemsep}{0mm} 
     \setlength{\labelsep}{0.5em} 
     \setlength{\itemindent}{0mm} 
     \setlength{\listparindent}{6mm} 
    }}
{\end{list}}

\title{Learning Transferable Conceptual Prototypes for Interpretable Unsupervised Domain Adaptation}
\author{Junyu~Gao, Xinhong Ma,
	and~Changsheng~Xu,~\IEEEmembership{Fellow,~IEEE}
	\IEEEcompsocitemizethanks{	
	Junyu Gao, Xinhong Ma, and Changsheng Xu are with the State Key Laboratory of Multimodal Artificial Intelligence Systems (MAIS), Institute of Automation, Chinese Academy of Sciences, Beijing 100190, P. R. China, and with School of Artificial Intelligence, University of Chinese Academy of Sciences, Beijing, China.  Changsheng Xu is also with the PengCheng Laboratory, Shenzhen 518066, China. (e-mail: junyu.gao@nlpr.ia.ac.cn; stefanxinhong@gmail.com; csxu@nlpr.ia.ac.cn).}
	
	\thanks{Copyright (c) 2021 IEEE. Personal use of this material is permitted. However, permission to use this material for any other purposes must be obtained from the IEEE by sending a request to pubs-permissions@ieee.org.}
}

\markboth{IEEE Transactions on Image Processing,~Vol. XX, ~No. XX,~Aug 201X}
{GAO \MakeLowercase{\textit{et al.}}: Learning Transferable Conceptual Prototypes for Interpretable Unsupervised Domain Adaptation}

\maketitle
\begin{abstract}
	\justifying Despite the great progress of unsupervised domain adaptation (UDA) with the deep neural networks, current UDA models are opaque and cannot provide promising explanations, limiting their applications in the scenarios that require safe and controllable model decisions. At present, a surge of work focuses on designing deep interpretable methods with adequate data annotations and only a few methods consider the distributional shift problem. Most existing interpretable UDA methods are post-hoc ones, which cannot facilitate the model learning process for performance enhancement. In this paper, we propose an inherently interpretable method, named Transferable Conceptual Prototype Learning (TCPL), which could simultaneously interpret and improve the processes of knowledge transfer and decision-making in UDA. To achieve this goal, we design a hierarchically prototypical module that transfers categorical basic concepts from the source domain to the target domain and learns domain-shared prototypes for explaining the underlying reasoning process. With the learned transferable prototypes, a self-predictive consistent pseudo-label strategy that fuses confidence, predictions, and prototype information, is designed for selecting suitable target samples for pseudo annotations and gradually narrowing down the domain gap. Comprehensive experiments show that the proposed method can not only provide effective and intuitive explanations but also outperform previous state-of-the-arts.

\end{abstract}


\section{Introduction}

Unsupervised domain adaptation (UDA) aims to transfer knowledge from a source domain with rich supervision to unlabeled domains. The mainstream solution of unsupervised domain adaptation is to reduce the shift of the source and target data distributions, a.k.a. \emph{domain alignment}. Among existing UDA methods, deep learning-based approaches have made significant progress~\cite{cai2019learning,cui2020towards,liu2021cycle,chen2022reusing,rangwani2022closer}. However, we cannot completely trust the results produced by these black-box models~\cite{zeiler2014visualizing,zintgraf2017visualizing,zhou2016learning,gao2017deep}, 
especially in high-risk fields such as self-driving cars, diagnosis of cancer, etc. 

\begin{figure}[!t]
	\centering
	\includegraphics[width=1\linewidth]{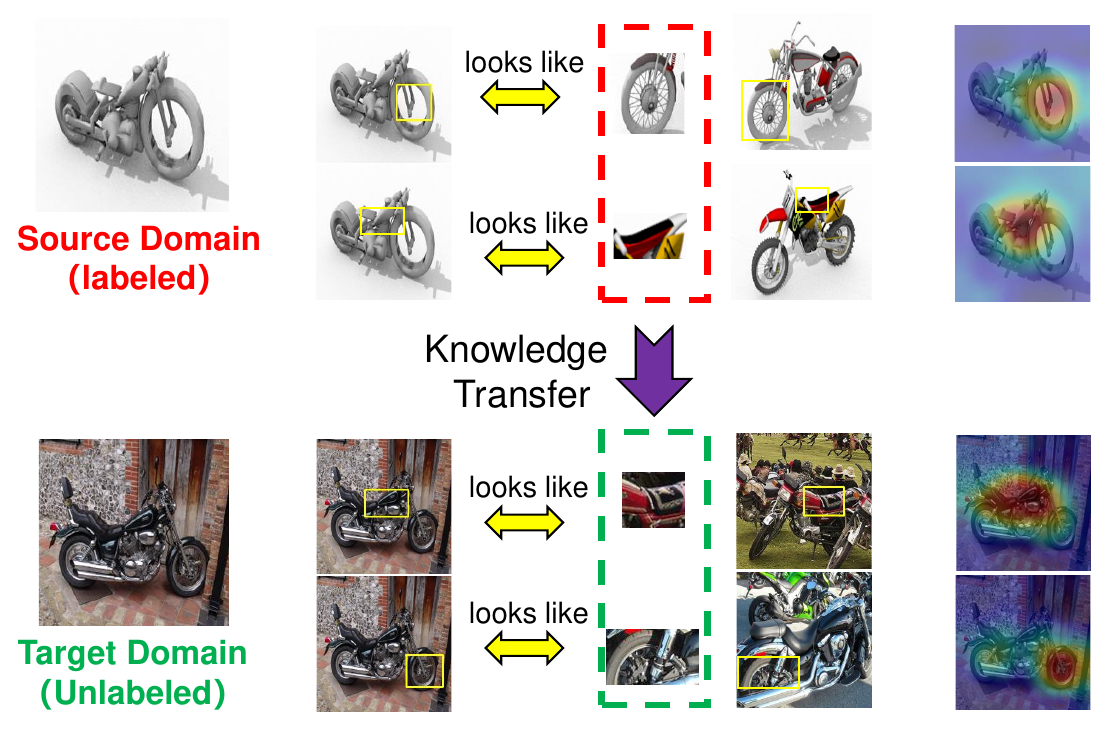}
	\caption{The illustrations of transferring categorical basic concepts for interpretable UDA.
		To recognize  \emph{motorcycle} images in the unlabeled target domain with interpretations, 
		basic concepts of \emph{motorcycle} category, \eg, wheels, and frame, are first learned in the labeled source domain for,
		and then transferred to the unlabeled target domain. 
		The visualizations (right column) highlight image patches mostly related to basic concepts of \emph{motorcycle}, which provides the clues why the given image is recognized.}
	\label{fig:motivation}

\end{figure}

In recent years, there has been a surge of work in discovering how a deep neural network processes input images and makes predictions \cite{gilpin2018explaining,selvaraju2017grad,zhou2016learning}.
Some methods explore post-hoc interpretability which explains the decision making process of trained models by visualizing the correlation between input pixels (or low-level features) and the final outputs~\cite{ribeiro2016model,zhou2016learning,selvaraju2017grad,wang2020score}.
However, these methods cannot offer guidance on how to correct mistakes made by the original model for performance improvement.
The other interpretable learning strategy directly designs a model to learn representations that agree with human thinking in making predictions~\cite{chen2019looks,ge2021peek,wang2021interpretable,liang2020training}.
%
Despite promising interpretation, these methods require elaborate  design, and their performance is not competitive with non-interpretable methods.
Besides, all the above interpretable methods are designed for the scenarios with adequate data annotations and few of them can handle the distribution shift between domains. Although some pioneering works~\cite{hou2021visualizing,zunino2021explainable,petryk2022guiding} explore interpretable learning paradigms for tackling the domain distribution shift problem, they are still post-hoc methods~\cite{hou2021visualizing,zunino2021explainable} or require external language prior~\cite{petryk2022guiding}. Moreover, it is still unclear what visual knowledge is transferred from the source domain, and how does the transferred knowledge influence model decisions when facing short of labeled data in the target domain.

In this paper, we argue that designing a reliable and inherently interpretable UDA model should consider the following two questions: (1) \emph{What knowledge should be transferred from the source domain to the target domain?} and (2) \emph{How to ensure that the explainable information from the source domain could be reliably transferred to the unlabeled target domain and effectively improve target performance?} To address the questions, we propose to transfer categorical basic concepts for interpretable unsupervised domain adaptation.
Considering the example in~\figref{motivation},
to recognize a object in a (new) target domain, we humans may find that the wheels and frame in the image look like those of \emph{motorcycle} in the source domain. In fact, humans usually explain their reasoning process by dissecting the image into object parts and pointing out the evidence from these identified parts to
the concepts stored in his / her mind (source domain), and make a final decision~\cite{tenenbaum2011grow,stach2004local}. Therefore, it is necessary to learn representations of basic concepts for each category, and then calculate the similarity to test images in order to implement interpretable object recognition. Since the target domain lacks labeled data, it is difficult to learn robust categorical basic concepts, motivating us to transfer source basic concepts of \emph{motorcycle} to help recognize \emph{motorcycle} images in the target domain.

Based on the above observations, in this paper, we propose a Transferable Conceptual Prototype Learning (TCPL) method, 
which aims to learn domain-shared conceptual prototypes for transferring explainable categorical knowledge and explaining the underlying decision reasoning process for UDA.
Specifically,
we first propose a Hierarchically Prototypical Module (HPM), 
which bridges high-level features with a set of basic conceptual prototypes so that
the model's predictions could be disentangled and visualized for interpretable classification. A hierarchical structure is adopted to extract multigrained feature maps for each image, making sure that the conceptual meanings represented by prototypes are consistent at different spatial scales.
To encode categorical conceptual information into prototypes, a prototype transparency strategy is further designed.
Finally, to enhance the transferability of prototypes, we design a self-predictive consistent pseudo-label strategy, which leverages the clues of classification confidences, predictions, and prototypes to mine reliable pseudo labels for target samples, promoting the domain-shared prototype learning and reducing the domain gap.

Our contributions can be summarized as follows:
(1) Targeting at interpretable unsupervised domain adaptation, we propose a transferable conceptual prototype learning (TCPL) method,
which transfers categorical basic concepts to the target domain and learns domain-shared prototypes for explaining the underlying reasoning process.
(2) To learn reliably transferable conceptual prototypes for decision explanation, we propose a hierarchically prototypical module learned with an interpretable prototype learning strategy and a self-predictive consistent pseudo-label strategy. 
(3) Extensive experiments on three benchmarks verify that the proposed method can not only provide effective and intuitive explanations but also outperform previous state-of-the-arts by a considerable margin.

\section{Related Work}

\noindent{\textbf{Unsupervised Domain Adaptation (UDA).}}
UDA adopts a model trained on a labeled source domain to an unlabeled target domain.
Some UDA methods learn domain-invariant features via minimizing the discrepancy between domains \cite{long2015learning,long2017deep,sun2016deep,liu2021cycle,li2021implicit,hu2022learning,wang2022probability}, 
\eg, Tzeng \etal ~\cite{tzeng2014deep} introduce an adaptation layer and a domain confusing loss to learn semantically meaningful and domain-invariant representations.
Inspired by generative adversarial networks (GANs) \cite{goodfellow2014generative},
another family of UDA methods applies adversarial learning to obtain domain-invariant representations \cite{ganin2015unsupervised,li2019joint,long2018conditional,rangwani2022closer,dai2021disentangling}.
Despite the success achieved by domain alignment, class discrimination also loses due to the distorted structure of semantic features \cite{cai2019learning,tang2020unsupervised}.
How to maintain class discriminability has also been considered by recent UDA work \cite{cui2020towards,li2020enhanced,li2020domain,pan2019transferrable,tang2020unsupervised,chen2022reusing,deng2021joint}. 
Despite superior performance,
the above deep unsupervised domain adaptation methods are still black-box models.
To solve this problem, this paper focuses on exploring reliable interpretation for deep unsupervised domain adaptation models.

\begin{figure*}[!t]
    \centering
    \includegraphics[width=0.95\linewidth]{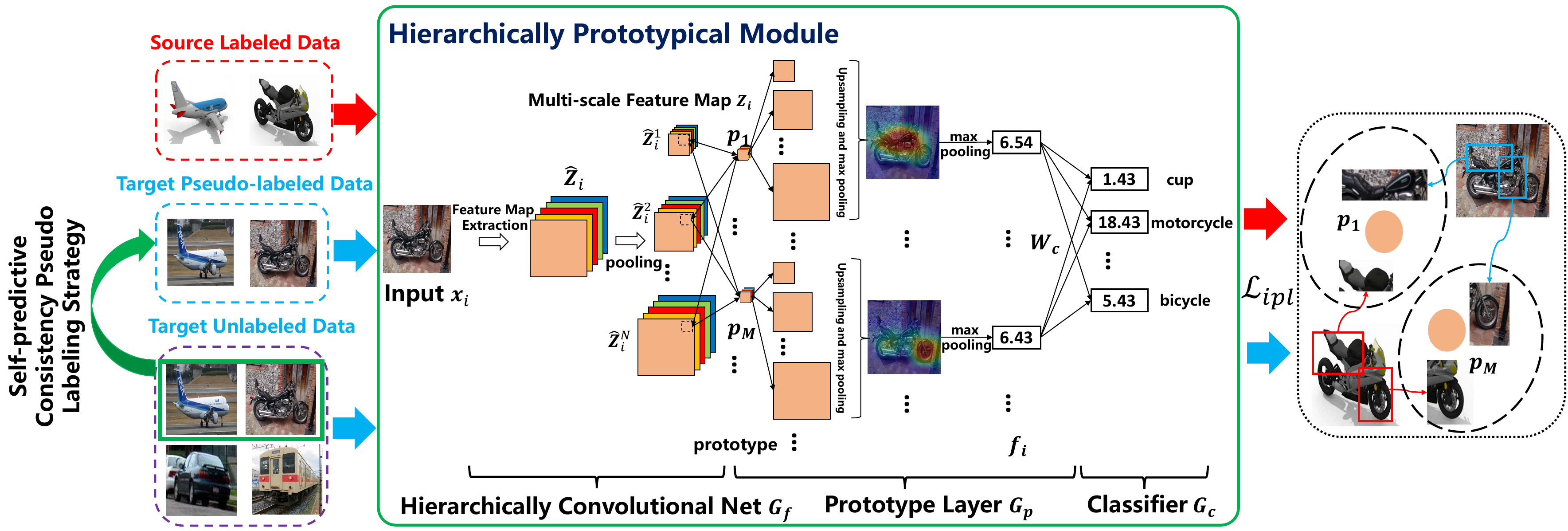}
    \caption{The overview of transferable conceptual prototype learning (TCPL) for interpretable UDA.
    The hierarchically prototypical module captures multi-scale information within images and encodes categorical basic concepts into prototypes for interpretable classification.
    The self-predictive consistent pseudo labeling strategy assigns high-confident pseudo labels to target samples for enhancing the transferability of the prototype.
    TCPL is learned in an end-to-end manner by minimizing the interpretable prototype learning loss ${\mathcal L}_{ipl}$.
    }
    \label{fig:model}

\end{figure*}

\noindent{\textbf{Interpreting Neural Networks.}}
The substantial applications of deep learning have necessitated the development of interpretable methods for neural networks,
which can be mainly divided into post-hoc interpretable methods and interpretable representation learning methods.
The post-hoc interpretable methods focus on exploring interpretability of trained neural networks
by analyzing  perturbations of input images \cite{zeiler2014visualizing,zhou2015predicting,zintgraf2017visualizing,petsiuk2018rise,ribeiro2016should}, 
salience maps \cite{selvaraju2017grad,chattopadhay2018grad,zhou2016learning,fukui2019attention,li2018tell,zheng2019re,wang2019sharpen}, etc. 
%
%
However, the resulting explanations of the above methods are insufficient to provide in-depth reasoning for model inference and cannot improve model performance during training. The other line of interpreting neural networks designs inherently interpretable models via learning interpretable representations \cite{chen2019looks,ge2021peek,wang2021interpretable,ghorbani2019towards},
which handle with high-level convolutional layers to extract more human-intuitive concept-level explanations.
For example,
some work \cite{bau2017network,gonzalez2018semantic,zhang2018interpretable,liang2020training} encourage each filter to respond to a specific concept by adding regularization terms.
%
Despite satisfying interpretation, the performance of these methods is not competitive with non-interpretable methods.
Besides, all the above models are designed for the scenarios with adequate data annotations
and none of them consider the distribution shift problems.

\noindent{\textbf{Interpretability with Domain Gap.}}
To the best of our knowledge, only a few works explore the interpretability of neural networks for the domain gap problem.
Hou \etal ~\cite{hou2021visualizing} try to understand the adaptation process by visualizing domain knowledge differences with image translation.
Zunino \etal ~\cite{zunino2021explainable} enforces
a periodic saliency-based feedback to encourage the model
to focus on relevant image regions.
However, the two methods are still post-hoc methods. Although~\cite{zhang2022explaining} attempt to learn the nearest source samples of a target sample as reference upon which the classifier makes the decision, it ignores the intrinsic local-to-global relations in object recognition and cannot further improve the model performance. Petryk \etal~\cite{petryk2022guiding} use a high-level language specification as guidance for constraining the classification evidence to task-relevant features, instead of distractors. However, this method is highly depend on large-scale multimodal models, such as CLIP~\cite{radford2021learning}.
Different from the above approaches, we propose to learn transferable conceptual prototypes in a pure vision-driven manner to explain the process of both knowledge transfer and model inference. 

\section{Our Approach}

%
In unsupervised domain adaptation, there are ${n_s}$ labeled samples $\left\{ {\left( {{\mathbf x}_i^s, y_i^s} \right)} \right\}_{i = 1}^{{n_s}}$ from the source domain ${{\mathcal D}_S}$, 
where ${\mathbf x}_i^s \in {{\mathcal X}_S}$, $y_i^s \in {{\mathcal Y}_S}, {{\mathcal Y}_S}=\left\{ {1,2,\cdots,c,\cdots ,C} \right\}$.
$\mathcal{X_S}$ and ${\mathcal Y}_S$ are defined as the source data space and source label space, respectively.
Additionally,  ${n_t}$ unlabeled samples $\left\{ {\left({\mathbf x}_i^t  \right)} \right\}_{i = 1}^{{n_t}}$ from the target domain ${{\mathcal D}_T}$ are also given, 
where ${\mathbf x}_i^t \in {{\mathcal X}_T}$,  and the ${{\mathcal X}_T}$ represents target data space. 
The ${{\mathcal X}_S}$ and ${{\mathcal X}_T}$ are assumed to be different but related (referred as domain shift~\cite{shimodaira2000improving}). 
The target task is assumed to be the same with the source task, \ie, the source label space ${\mathcal Y}_S$ is shared with the target label space ${\mathcal Y}_T$. 
Our goal is to develop a deep neural network $f$: ${{\mathcal X}_T} \to {{\mathcal Y}_T}$ that is able to predict labels for samples from target domain,
and interpret the process of knowledge transfer and model inference.

As shown in~\figref{model}, the proposed Transferable Conceptual Prototype Learning (TCPL) method aims to explain the underlying decision reasoning process for interpretable unsupervised domain adaptation.
The core architecture of TCPL is the Hierarchically Prototypical Module (HPM), 
which inputs source labeled data ${\mathcal D}_S$, target unlabeled data ${\mathcal D}_T$ and target pseudo-labeled data ${\mathcal D}_{PLT}$,
and aims to learn domain-shared prototypes for interpretable classification. 
Specifically, 
HPM leverages the supervision from source labeled data ${\mathcal D}_S$ and target pseudo-labeled data ${\mathcal D}_{PLT}$ to minimize interpretable prototype learning loss ${\mathcal L}_{ipl}$,
which helps to capture categorical concepts information and maintain prototype learning. 
To transfer interpretable reasoning knowledge from the source domain to the unlabeled target domain,
we design a self-predictive consistent pseudo labeling strategy that fuses confidence, predictions, and prototype information to 
construct target pseudo-labeled dataset ${\mathcal D}_{PLT}$ and assist domain-shared prototype learning.
In the following sections, we first introduce the workflow of the hierarchically prototypical module.
Then, we present how to learn semantic prototypes for interpretable classification,
and how to enhance the transferability of prototypes.

\subsection{Hierarchically Prototypical Module}\label{sec:hpm}

To implement interpretable classification, 
the hierarchically prototypical module (HPM) consists of three modules:
hierarchically convolutional net $G_f$ with parameters $\theta_f$,
prototype layer $G_p$ with prototypes $\mathbf{P}$,
and a classifier $G_c$ with the weight matrix $\mathbf{W}_c$, as shown in Figure \ref{fig:model}.
HPM inputs source labeled data ${\mathcal D}_S$, target unlabeled data ${\mathcal D}_T$, and target pseudo-labeled data ${\mathcal D}_{PLT}$.

$G_f$ takes traditional network pretrained on the ImageNet \cite{ILSVRC15} as the backbone, \eg, ResNet-50~\cite{he2016deep},  
and extra $1\times1$ convolutional layers are added to adjust the number of channels for top-level feature map.
%
%
Specifically, 
given an input image ${\mathbf x}_i$,
the feature map $\mathbf{\hat Z}_i \in \mathbb{R}^{W \times H \times D}$ is extracted by the backbone with spatial resolution $W \times H$ and $D$ channels. Note that for convenience, the domain marks $s$ and $t$ are removed in this subsection.
Afterward, to obtain fruitful information at different scales, we simply utilize multiple max-pooling layers to operate upon $\mathbf{\hat Z}_i$.
The sizes of the pooling operation are $\left\{k_n(\cdot)\right\}_{n=1}^N$, where $N$ is the number of the pooling layers.
That is,
the rectangular pooling region with the size $k_n(\cdot)$ at each location of $\mathbf{\hat Z}_i$ is down-sampled to the max value of the region, 
resulting in the multi-scale pooled feature maps $\mathbf{Z}_i = \left\{ \mathbf{\hat Z}_i^1, \cdots, \mathbf{\hat Z}_i^n, \cdots, \mathbf{\hat Z}_i^N \right\}$.
In this way, each $\mathbf{\hat Z}_i^n \in \mathbb{R}^{W_n \times H_n \times D}$  can encode semantic information for a specific image scale and layout.
We denote the above process as $\mathbf{Z}_i=G_f({\mathbf x}_i)$.

Once $\mathbf{Z}_i$ is available, the prototype layer $G_p$ will project it into the embedding space spanned by learnable prototypes.
Specifically, $G_p$ learns $M$ prototypes $\mathbf{P}^{(c)}=$ $\left\{\mathbf{p}_{j}^{(c)}\right\}_{j=1}^{M}$ for each category,
and there are a total of $M \times C$ prototypes  $\mathbf{P}=\left\{\left\{\mathbf{p}_{j}^{(1)}\right\}_{j=1}^{M}, \cdots,\left\{\mathbf{p}_{j}^{(C)}\right\}_{j=1}^{M}\right\}$  for $C$ categories.
For convenience, prototypes are denoted as $\mathbf{P}=\left\{{\mathbf p}_j\right\}_{j=1}^{M \times C}$, where ${\mathbf p}_j \in \mathbb{R}^{W_p \times H_p \times D}, \left\| {\mathbf p}_j \right\|=1$,
where $W_p=H_p=1$ and the channel of prototype is the same as $\mathbf{Z}_i$.
Each prototype is used to represent some prototypical activation patterns in a grid of each $\mathbf{\hat Z}_i^n$, which in turn correspond to some prototypical image patches in the original pixel space.
For example, as shown in Figure \ref{fig:model},
the prototype can be understood as the basic conceptual information of representative parts of motorcycle images. 
%
%

In our pipeline, given the output of the hierarchically convolutional net $\mathbf{Z}_i$,
each basis unit ${\mathbf p}_j$ in the prototype layer $G_p$ computes the similarity from all $1 \times 1$ grids of $\mathbf{Z}_i$ to the $j$-th prototype.
This similarity map keeps the size of the pyramid feature map and retains the spatial relation well, which can be upsampled to the original image size to determine image regions similar to the $j$-th prototype.
Then the compact similarity map is reduced to one value by
global max pooling, \ie, $G_{p_j}(\mathbf{Z}_i)=\mathop {\max }\nolimits_{{\mathbf b} \in \operatorname{grids}\left(\mathbf{Z}_i\right)}{\mathbf b}^{\top}{\mathbf p}_j$,
which is the maximum probability that the semantic concepts represented by prototype ${\mathbf p}_j$ appear in the current image.  
Finally, the projection probability on all basis units ${\mathbf f}_i = G_p(\mathbf{Z}_i) = \left[G_{p_1}(\mathbf{Z}_i),\cdots, G_{p_{M \times C}}(\mathbf{Z}_i)  \right], {\mathbf f}_i \in \mathbb{R}^{(M \times C)}$
are taken as the input of classifier $G_c$ which is composed of
a fully-connection layer with the weight matrix $\mathbf{W}_c$. 
The final label is predicted according to the logistic regression.

To encode basic conceptual information of different categories into prototypes,
we design an interpretable prototype learning loss ${\mathcal L}_{ipl}$.
Also, we use the learned prototypes as queries to search for similar image block features in the dataset and further update prototypes,
so that all prototypes can be traced back to conceptual patches in the datasets for interpretability. Details are in the following subsections.

\subsection{Learning Conceptual Prototypes}

The proposed method aims to learn a group of conceptual prototypes, 
which can be used for predicting categorical labels and visualizing the reasoning evidence.
To learn discriminative prototypes with specific conceptual meanings,
three requirements should be taken into consideration:
(1) The representations of prototypical image patches from the same category should be clustered around the corresponding prototypes;
(2) Prototypes of different categories are far away from each other so that features with different conceptual meanings are disentangled;
(3) Model inference should rely on the learned prototypes, making it possible to interpret the reasoning process via visualizing decision evidence.
To achieve the above goals, 
we propose an interpretable prototype learning strategy including the interpretable prototype learning loss ${\mathcal L}_{ipl}$ and a prototype transparency strategy.
The former encourages the model to capture categorial concepts from datasets for discriminative prototype learning.
The latter updates prototypes with the semantic-related image patches in the datasets so that the classification process can be visualized with conceptual image patches for visual interpretability.

\noindent{\textbf{Interpretable Prototype Learning.}}
The overall objective of interpretable prototype learning ${\mathcal L}_{ipl}$ is shown as follows:
\begin{equation}\label{eq:mainp}
{\mathcal L}_{ipl} =  {\mathcal L}_c  + \lambda_{1} {\mathcal L}_{\text{cdpd}} + \lambda_{2} {\mathcal L}_{\text{dd}},
\end{equation}
where ${\mathcal L}_c$ is a cross-entropy loss on source labeled data ${\mathcal D}_S$ and target pseudo-labeled data ${\mathcal D}_{PLT}$.
The cross-domain prototype discrimination loss ensures that the representation of each prototypical region within a training image is pushed to one of the prototypes belonging to the ground-truth category. It should also push the representation of prototypical image regions away from prototypes of other categories. 
%
%
The decision disentangled loss ${\mathcal L}_{\text{dd}}$ decouples predictions of different categories,
making sure that the classification score of each category is only decided by prototypes of the corresponding category and has nothing to do with others.
$\lambda_{1}, \lambda_{2}$ are hyper-parameters to balance the corresponding terms.
The details are as follows.

\textbf{Cross-domain prototype discrimination loss:}
The prototype layer $G_p$ projects image patches to the embedding spaces spanned by prototypes
and preserves the essential categorical information for interpretable classification.
Therefore, the operation should encourage each training image to have some latent patch that is close to at least one prototype of its category
and stay away from the prototypes of other categories.
The above requirements are achieved by the proposed cross-domain prototype discrimination loss:

\begin{equation}\label{eq:cdpd}
	\mathcal{L}_{\text {cdpd}} = \frac{\exp{({\widetilde{r}}^s}+\eta{\widetilde{r}}^t)}{\exp{(r^s+\eta r^t)}}, 
\end{equation}
where $r^s = {\mathbb E}_{\left({\mathbf x}_i, y_i\right) \in {\mathcal D}_S } \min _{\mathbf{p}_{j}    \in \mathbf{p}^{(y_{i})},\mathbf{b} \in \operatorname{grids}\left({\mathbf Z}_i\right)} \frac{\mathbf{b}^{\top} \mathbf{p}_{j}}{\|\mathbf{b}\|}$, and
${\widetilde{r}}^s = {\mathbb E}_{\left({\mathbf x}_i, y_i\right) \in {\mathcal D}_S } \min _{\mathbf{p}_{j}    \notin \mathbf{p}^{(y_{i})},\mathbf{b} \in \operatorname{grids}\left({\mathbf Z}_i\right)} \frac{\mathbf{b}^{\top} \mathbf{p}_{j}}{\|\mathbf{b}\|}$. $r^t$ and ${\widetilde{r}}^t$ can be calculated similarly by using the target pseudo-labeled data ${\mathcal D}_{PLT}$. $\eta$ is a balanced term for controlling the prototype preference between source and target domains. Here, given the prototypes $\mathbf{P}=\left\{{\mathbf p}_j\right\}_{j=1}^{M \times C}$ and feature maps $\left\{ {\mathbf Z}_i | {\mathbf Z}_i=G_f({\mathbf x}_i) \right\} $, $\mathbf{p}^{(y_{i})}$ represents the collection of prototypes belonging to category $y_{i}$.
The operation $\operatorname{grids}\left(\cdot\right)$ is a $1 \times 1$ spatial sampling operation on the feature maps, which remains channels unchanged.
%
With this objective,
semantic similar latent patches from source or target domains gather around their conceptual prototypes and stay away from other prototypes,
making it possible for learning a discriminative and domain-shared feature space to interpret model reasoning.

\textbf{Decision disentangled loss:}
Once obtaining the feature space spanned by prototypes,
we can effectively build classifier $G_c$ by optimizing the weight matrix $\mathbf{W}_c$.
%
Different from traditional classifier based on fully-connected layer,
it is expected that $\mathbf{W}_{(c, j)} \approx 0 $  (initially fixed at -0.5) 
if the $j$-th conceptual prototype (corresponds to $\mathbf{p}_{j}$) does not belong to the $c$-th class, and each class is only related to its own prototypes. 
Thus, the sparse constraint on the weight matrix, termed as decision disentangled loss ${\mathcal L}_{\text{dd}}$, is represented as follows.
\begin{equation}\label{eq:dd}
{\mathcal L}_{\text{dd}} =  \sum_{c=1}^{C} \sum_{j: \mathbf{p}_{j} \notin \mathbf{p}^{(c)}}\text{smoothL1}\left(\mathbf{W}_{(c, j)}\right),
\end{equation}
where $\mathbf{W}_{(c, j)}$ is the connection weight of the $j$-th prototype to the $c$-th class. $\text{smoothL1} (\cdot)$~\cite{girshick2015fast} is the modified version of the L1 loss.
The decision disentangled loss guarantees that the discriminative evidence comes from the prototypes of the ground truth class as much as
possible and relies less on the prototypes of negative classes.
Besides, the contribution of prototypes to the final prediction can be quantified by the weight matrix $\mathbf{W}_c$ for decision explanation.

\begin{algorithm}[!t]
	\small
	\caption{\small The proposed TCPL}
	\label{alg:train}
	\textbf{Input}: source labeled dataset ${\mathcal D}_S=\left\{ {\left({\mathbf x}_i^s, y_i^s  \right)} \right\}_{i = 1}^{{n_s}}$,
	target unlabeled dataset ${\mathcal D}_T = \left\{ {\left({\mathbf x}_i^t  \right)} \right\}_{i = 1}^{{n_t}}$,
	target pseudo-labeled dataset ${\mathcal D}_{PLT} = \emptyset$. \\
	\textbf{Initialization}: The backbone of $G_f$ is a pre-trained model on ImageNet and the rest  of $G_f$ and prototypes are randomly initialized;
	classifier $G_c$ follows the criteria that $\mathbf{W}_{(c, j)}=1$ if ${\mathbf p}_j \in {\mathbf p}^{(c)}$, otherwise $\mathbf{W}_{(c, j)}=-0.5$.\\
	\For{$epoch=1$ to $epoch\_num$}
	{
		Optimizing paramters of hierarchically prototypical module $\theta_f, \theta_p, \theta_c$ via Eq \eqref{eq:mainp}. \\
		Updating target pseudo-labeled dataset ${\mathcal D}_{PLT}$ with the self-pretictive consisitency pseudo label strategy.\\
		\If{$epoch > epoch\_update\_proto$}
		{
			Updating prototypes via Eq \eqref{eq:psl}.
		}
	}
\end{algorithm}

\noindent{\textbf{Prototype Transparency.}}
To visualize the prototypes as training image patches,
we push each prototype $\mathbf{p}_{j}$ onto the nearest latent training patch from the same class.
In this way,
each basis prototype can be traced back its nearest image patches from the same class,
and we can conceptually equate each prototype with a training image patch.
Thus, users can explicitly know the conceptual meaning of each prototype.
To achieve this goal, for prototype $\mathbf{p}_{j}$ and features ${\mathbf Z}_i$ of class $c$, we perform the following update:
%
\begin{equation}\label{eq:psl}
	\mathbf{p}_{j} \leftarrow \arg \max_{\mathbf{b}^s~ \text{or} ~\mathbf{b}^t} \max\left(\frac{{\mathbf{b}^s}^{\top} \mathbf{p}_{j}}{\|\mathbf{b}^s\|},~ \eta\frac{{\mathbf{b}^t}^{\top} \mathbf{p}_{j}}{\|\mathbf{b}^t\|}\right),
\end{equation}
where $\mathbf{b}^s \in \{{\mathbf Z}^s_i\}$, which is the extracted grid feature from the source data ${\mathcal D}_S$. Similarly, $\mathbf{b}^t$ is from the target pseudo labeled data ${\mathcal D}_{PLT}$. The above equation means that we select the most similar patches from either source of target pseudo labeled images jointly with the balanced term $\eta$.
%
It is noticed that the prototype update is added into the training when the model approaches convergence.
If this strategy is added in the early of training, the prototypes change greatly, damaging the stability of model training.
%

\begin{table*}[htbp]
	\centering
	\caption{Classification Accuracy (\%) on Office-Home dataset. The best results are {\color{red}{\textbf{red}}} and the second-best results are \underline{underlined}.}
	\small
	\scalebox{0.9}{
		\renewcommand{\arraystretch}{1.1}
		\begin{tabular}{cccccccccccccc}
			\toprule
			Source & Ar    & Ar    & Ar    & Cl    & Cl    & Cl    & Pr    & Pr    & Pr    & Rw    & Rw    & Rw    & \multirow{2}[0]{*}{AVG} \\
			Target & Cl    & Pr    & Rw    & Ar    & Pr    & Rw    & Ar    & Cl    & Rw    & Ar    & Cl    & Pr    &  \\
			\midrule
			ResNet-50 & 34.90  & 50.00  & 58.00  & 37.40  & 41.90  & 46.20  & 38.50  & 31.20  & 60.40  & 53.90  & 41.20  & 59.90  & 46.10  \\
			DANN  & 45.60  & 59.30  & 70.10  & 47.00  & 58.50  & 60.90  & 46.10  & 43.70  & 68.50  & 63.20  & 51.80  & 76.80  & 57.60  \\
			CDAN  & 50.70  & 70.60  & 76.00  & 57.60  & 70.00  & 70.00  & 57.40  & 50.90  & 77.30  & 70.90  & 56.70  & 81.60  & 65.80  \\
			CDAN+VAT+Entropy & 52.20  & 71.50  & 76.40  & 61.10  & 70.30  & 67.80  & 59.50  & 54.40  & 78.60  & 73.20  & 59.00  & 82.70  & 67.30  \\
			FixMatch & 51.80  & 74.20  & 80.10  & 63.50  & 73.80  & 61.30  & 64.70  & 51.40  & 80.00  & 73.30  & 56.80  & 81.70  & 67.70  \\
			MDD   & 54.90  & 73.70  & 77.80  & 60.00  & 71.40  & 71.80  & 61.20  & 53.60  & 78.10  & 72.50  & 60.20  & 82.30  & 68.10  \\
			MDD+IA & 56.20  & 77.90  & 79.20  & 64.40  & 73.10  & 74.40  & 64.20  & 54.20  & 79.90  & 71.20  & 58.10  & 83.10  & 69.50  \\
			GSDA  & 61.30  & 76.10  & 79.40  & 65.40  & 73.30  & 74.30  & 65.00  & 53.20  & 80.00  & 72.20  & 60.60  & 83.10  & 70.30  \\
			GVB-GD & 57.00  & 74.70  & 79.80  & 64.60  & 74.10  & 74.60  & 65.20  & 55.10  & 81.00  & 74.60  & 59.70  & 84.30  & 70.40  \\
			UTEP   & 57.40  & 76.10  & 80.20  & 64.20  & 73.20  & 73.70  & 64.80  & 55.40  & 80.90  & 74.70  & 61.10  & 84.60  & 70.60  \\ 
			RSDA-MSTN & 53.20  & 77.70  & 81.30  & 66.40  & 74.00  & 76.50  & 67.90  & 53.00  & 82.00  & 75.80  & 57.80  & 85.40  & 70.90  \\
			SRDC  & 52.30  & 76.30  & 81.00  & \color{red}{\textbf{69.50}}  & 76.20  & \underline{78.00}  & 68.70  & 53.80  & 81.70  & 76.30  & 57.10  & 85.00  & 71.30  \\
			DALN   & 57.80  & \underline{79.90}  & 82.0  & 66.30  & 76.20  & 77.20  & 66.70  & 55.50  & 81.30  & 73.00  & 60.40  & 85.30  & 71.80  \\ 
			SENTRAY & \color{red}{\textbf{61.80}}  & 77.40  & 80.10  & 66.30  & 71.60  & 74.70  & 66.80  & \color{red}{\textbf{63.00}}  & 80.90  & 74.00  & \color{red}{\textbf{66.30}}  & 84.10  & 72.20  \\
			SDAT   & 58.20  & 77.10  & 82.20  & 66.30  & \underline{77.60}  & 76.80  & 63.30  & 57.0  & 82.20  & 74.90  & 64.70  & 86.0  & 72.20  \\
			FixBi  & 58.10  & 77.30  & 80.40  & 67.70  & \color{red}{\textbf{79.50}}  & \color{red}{\textbf{78.10}}  & 65.80  & 57.90  & 81.70  & \underline{76.40}  & 62.90  & \underline{86.70}  & 72.70  \\
			CST   & 59.00  & 79.60  & \color{red}{\textbf{83.40}}  & 68.40  & 77.10  & 76.70  & \underline{68.90}  & 56.40  & \underline{83.00}  & 75.30  & 62.20  & 85.10  & \underline{73.00}  \\ \midrule
			TCPL  & \underline{61.20}  & \color{red}{\textbf{80.50}}  & \underline{82.80}  & \underline{68.80}  & 75.10  & 76.50  & \color{red}{\textbf{71.70}}  & \underline{59.80}  & \color{red}{\textbf{83.50}}  & \color{red}{\textbf{78.10}}  & \underline{66.20}  & \color{red}{\textbf{87.60}}  & \color{red}{\textbf{74.32}}  \\
			\bottomrule
		\end{tabular}%
	}
	\label{tab:officehome}%

\end{table*}%

\begin{table}[t!]
	\centering
	\caption{Average Class Accuracy  (\%) on VisDa dataset. The best results are  {\color{red}{\textbf{red}}}.}
	\begin{tabular}{crr}
		\toprule
		Method & \multicolumn{1}{c}{ResNet-50} & \multicolumn{1}{c}{ ResNet-101 } \\
		\midrule
		DANN  & 69.3  & 79.5 \\
		VAT   & 68.0 $\pm$ 0.3 & 73.4 $\pm$ 0.5 \\
		DIRT-T & 68.2 $\pm$ 0.3 & 77.2 $\pm$ 0.5 \\
		MCD   & 69.2  & 77.7 \\
		CDAN  & 70    & 80.1 \\
		CBST  & -     & 76.4 $\pm$ 0.9 \\
		KLD   & -     & 78.1 $\pm$ 0.2 \\
		MDD   & 74.6  & 81.6 $\pm$ 0.3 \\
		AFN   & -     & 76.1 \\
		STAR  & -     & 82.7 \\
		CDAN+VAT+Entropy & 76.5 $\pm$ 0.5 & 80.4 $\pm$ 0.7 \\
		MDD+IA & 75.8  & - \\
		MDD+FixMatch & 77.8 $\pm$ 0.3 & 82.4 $\pm$ 0.4 \\
		MixMatch & 69.3 $\pm$ 0.4 & 77.0 $\pm$ 0.5 \\
		FixMatch & 74.5 $\pm$ 0.2 & 79.5 $\pm$ 0.3 \\
		SENTRY & 76.7  & - \\
		CST   & 80.6 $\pm$ 0.5 & 86.5 $\pm$ 0.7 \\
		TCPL  & {\color{red}{\textbf{82.1}}} $\pm$ 0.2 & {\color{red}{\textbf{87.8}}} $\pm$ 0.5 \\
		\bottomrule
	\end{tabular}%
	\label{tab:visda}%
\end{table}%

\subsection{Self-predictive Consistent Pseudo Label}


Our method leverages conceptual prototypes for interpretable classification. Although in the above subsection, conceptual prototypes are jointly learned from the source and pseudo labeled target domains, the mining of reliable samples for pseudo labeling in ${\mathcal D}_T$ is not trivial. Therefore, we are motivated to leverage comprehensive information for pseudo label mining and transferring conceptual prototypes between the source and target domains.
Currently, most methods focus on designing robust pseudo-label metrics based on the predicted classification score or entropy to select high-confidence samples \cite{manders2018simple,zhang2018collaborative,kang2019contrastive,tan2020class,gao2019smart}. Due to the existence of domain gap, the model learned with source supervision is biased towards the source domain. Note that the previous approaches ignore the local (multi-scale) interpretable information for determining target pseudo labels, which is less discriminative and comprehensive.
%
%
To alleviate this issue,
we propose a robust self-predictive consistent pseudo label strategy, which uses predictive consistency under a committee of label-preserving image transformations as a more robust measure for sample selection.

For a target sample ${\mathbf x}_i^t$,
we first utilize classifier $G_c$ to predict its pseudo-label ${\bar y}_i^t = \arg \max G_c \circ G_p \circ G_f\left({\mathbf x}_i^t\right)$.
Then, we generate a committee of $q$ transformed version $\left\{Q_j({\mathbf x}_i^t)\right\}_{j=1}^q$ where $Q_j(\cdot)$ is an image transformation operation.
Finally, we calculate the classification confidence ${v}_{ij}^t$, pseudo label ${\bar y}_{ij}^t$ and the index of the most similar prototype ${w}_{ij}^t$ for each transformed instance $Q_j({\mathbf x}_i^t)$,
to evaluate the self-predictive consistency between the original image and transformed images:
\begin{equation}
    \begin{aligned}
    {v}_{ij}^t &= \max G_c \circ G_p \circ G_f\left(Q_j({\mathbf x}_i^t)\right)\\
    {\bar y}_{ij}^t &=\arg \max G_c \circ G_p \circ G_f\left(Q_j({\mathbf x}_i^t)\right) \\
    {w}_{ij}^t &=\arg \max G_p \circ G_f\left(Q_j({\mathbf x}_i^t)\right)
\end{aligned}
\end{equation}
Specifically,
the self-predictive consistency contains three criteria:
(1) classification confidence criteria: The classification confidences of the augmented samples are greater than a certain threshold, \ie, ${v}_{ij}^t > V$.
(2) prediction criteria: the model's classification result for a majority of augmented versions matches its prediction on the original image, \ie, $\operatorname{sum} \left({\bar y}_{ij}^t = {\bar y}_{i}^t\right) >  \frac{1}{2}q$.
(3) prototype criteria: the most similar prototype for a  majority of augmented versions is the same as the most similar prototype of the original image, \ie, $\operatorname{sum} \left({w}_{ij}^t \in \left\{{\bar y}_i^t \times N, \cdots, ({\bar y}_i^t +1 )\times N -1\right\}\right) > \frac{1}{2}q$.
Only if all three criteria are satisfied, we consider the sample as ``consistent'', which will be used for constructing target pseudo-labeled data $\mathcal{D}_{PLT}$ and safely join in the model training.
For instances marked as consistent, 
our approach minimizes the overall objective of interpretable prototype learning ${\mathcal L}_{ipl}$  to its augmented images rather than with respect to the original image itself,
which helps reduce overfitting and learns robust domain-shared prototypes. Algorithm \ref{alg:train} provides the training details of our full approach.
%


\section{Experiments}

We evaluate our on three popular benchmarks: Office-Home~\cite{venkateswara2017deep}, VisDA~\cite{8575439}, and DomainNet~\cite{peng2019moment}. 
Experiments show our effectiveness. 

\subsection{Experimental Setups}

\noindent{\textbf{Dataset.}}
We conduct experiments on three datasets. 
\textbf{Office-Home} \cite{venkateswara2017deep} is a challenging dataset, which consists of 15500 images from 65 categories. 
It is made up of 4 domains: Artistic (Ar), Clip-Art (CI), Product (Pr), and Real-World (Rw). 
\textbf{VisDA} \cite{8575439} is a large-scale dataset, where the source domain contains 15K synthetic images and the target domain consists of 5K images from the real world. 
\textbf{DomainNet} \cite{peng2019moment} is the largest domain adaptation dataset, which contains 600,000 images from six domains with 345 categories. 
The six domains are Clipart (clp), Infograph (inf), Painting (pnt), Quickdraw (qdr), Real (rel), and Sketch (skt), which can be used for constructing 30 transfer tasks.

\noindent{\textbf{Compared Methods.}}
We compare four types of baselines, namely feature adaptation methods, self-training methods, self-training methods for UDA, and other state-of-the-art UDA methods.
(1) Feature adaptation methods: DANN \cite{ganin2015unsupervised}, MCD \cite{saito2018maximum}, CDAN \cite{long2018conditional}, MDD \cite{zhang2019bridging}, MDD+IA \cite{jiang2020implicit},
BNM \cite{cui2020towards}, FixBi \cite{na2021fixbi}, CGDM \cite{du2021cross}, 
GSDA~\cite{hu2020unsupervised}, GVB-GD \cite{cui2020gradually},
SRDC \cite{tang2020unsupervised}, RSDA-MSTN \cite{gu2020spherical}, SWD \cite{lee2019sliced}, SDAT~\cite{rangwani2022closer}.
(2) Self-Training methods: We include VAT \cite{miyato2018virtual}, MixMatch \cite{berthelot2019mixmatch} and FixMatch \cite{sohn2020fixmatch}, in the semi-supervised learning literature as self-training methods. 
(3) Self-training methods for UDA: CBST \cite{Zou_2018_ECCV},  DIRT-T \cite{shu2018dirt}, KLD \cite{zou2019confidence}, CST \cite{liu2021cycle}, DALN~\cite{chen2022reusing}. 
We also create more powerful baselines: CDAN+VAT+Entropy (CDAVE) and MDD+Fixmatch (MDDF).
(4) Other SOTA: AFN \cite{xu2019larger}, STAR \cite{lu2020stochastic}, SENTRY \cite{prabhu2021sentry}, UTEP~\cite{hu2022learning}.

\noindent{\textbf{Implementation details.}}
The proposed method is implemented via Pytorch and applies the ResNet-50 pre-trained on ImageNet dataset as the backbone network. 
For the dataset VisDA, we add the experiments on the ResNet-101 backbone network to include more comparative methods. 
We utilize SGD optimizer for model learning.
The initial learning rate is 0.002, which is reduced to one-tenth every 50 epochs. To avoid the model overfitting issue on the source data, we follow~\cite{liu2021cycle} to leverage tsallis entropy  to facilitate the calibrations of target predictions.
The optimizer parameters of the tsallis loss follow \cite{liu2021cycle}. We follow~\cite{sohn2020fixmatch} to implement the image transformation operations $Q_j(\cdot)$.
The total training epoch is set as 250 and the prototype transparency begins at the 120-th epoch.
Hyper-parameters are tuned via cross-validation.
Specifically, we set $\lambda_1 = 0.88$, $\lambda_2 = 1e^{-4}$, $\eta=1$, and $V = 0.97$ in all transfer tasks. 
For each transfer task, we perform experiments three times 
and report the average classification accuracy.
All experiments run on a single NVIDIA GTX 3090 GPU.

%

%

\begin{table*}[htbp]
	\centering
	\caption{Classification Accuracy (\%) on DomainNet dataset. In each sub-table, the column-wise means source domain and the row-wise means target domain. The best results are  {\color{red}{\textbf{red}}}.}
	\scalebox{1}{
		\begin{tabular}{c|ccccccc||c|ccccccc}
			\toprule
			ResNet  & clp   & inf   & pnt   & qdr   & rel   & skt   & AVG   & CDAN  & clp   & inf   & pnt   & qdr   & rel   & skt   & AVG \\
			\midrule
			clp   & -     & 14.20  & 29.60  & 9.50  & 43.80  & 34.30  & 26.30  & clp   & -     & 13.50  & 28.30  & 9.30  & 43.80  & 30.20  & 25.00  \\
			inf   & 21.80  & -     & 23.20  & 2.30  & 40.60  & 20.80  & 21.70  & inf   & 18.90  & -     & 21.40  & 1.90  & 36.30  & 21.30  & 20.00  \\
			pnt   & 24.10  & 15.00  & -     & 4.60  & 45.00  & 29.00  & 23.50  & pnt   & 29.60  & 14.40  & -     & 4.10  & 45.20  & 27.40  & 24.20  \\
			qdr   & 12.20  & 1.50  & 4.90  & -     & 5.60  & 5.70  & 6.00  & qdr   & 11.80  & 1.20  & 4.00  & -     & 9.40  & 9.50  & 7.20  \\
			rel   & 32.10  & 17.00  & 36.70  & 3.60  & -     & 26.20  & 23.10  & rel   & 36.40  & 18.30  & 40.90  & 3.40  & -     & 24.60  & 24.70  \\
			skt   & 30.40  & 11.30  & 27.80  & 3.40  & 32.90  & -     & 21.20  & skt   & 38.20  & 14.70  & 33.90  & 7.00  & 36.60  & -     & 36.60  \\
			AVG   & 24.10  & 11.80  & 24.40  & 4.70  & 33.60  & 23.20  & 20.30  & AVG   & 27.00  & 12.40  & 25.70  & 5.10  & 34.30  & 22.60  & 21.20  \\
			\midrule
			SWD   & clp   & inf   & pnt   & qdr   & rel   & skt   & AVG   & BNM   & clp   & inf   & pnt   & qdr   & rel   & skt   & AVG \\
			\midrule
			clp   & -     & 14.70  & 31.90  & 10.10  & 45.30  & 36.50  & 27.70  & clp   & -     & 12.10  & 33.10  & 6.20  & 50.80  & 40.20  & 28.50  \\
			inf   & 27.70  & -     & 24.20  & 2.50  & 33.20  & 21.30  & 20.00  & inf   & 26.60  & -     & 28.50  & 2.40  & 38.50  & 18.10  & 22.80  \\
			pnt   & 33.60  & 15.30  & -     & 4.40  & 46.10  & 30.70  & 26.00  & pnt   & 39.90  & 12.20  & -     & 3.40  & 54.50  & 36.20  & 29.20  \\
			qdr   & 15.50  & 2.20  & 6.40  & -     & 11.10  & 10.20  & 9.10  & qdr   & 17.80  & 1.00  & 3.60  & -     & 9.20  & 8.30  & 8.00  \\
			rel   & 41.20  & 18.10  & 44.20  & 4.60  & -     & 31.60  & 27.90  & rel   & 48.60  & 13.20  & 49.70  & 3.60  & -     & 33.90  & 29.80  \\
			skt   & 44.20  & 15.20  & 37.30  & 10.30  & 44.70  & -     & 30.30  & skt   & 54.90  & 12.80  & 42.30  & 5.40  & 51.30  & -     & 33.30  \\
			AVG   & 31.50  & 13.10  & 28.80  & 6.40  & 36.10  & 26.10  & 23.60  & AVG   & 37.60  & 10.30  & 31.40  & 4.20  & 40.90  & 27.30  & 25.30  \\
			\midrule
			CGDM  & clp   & inf   & pnt   & qdr   & rel   & skt   & AVG   & TCPL  & clp   & inf   & pnt   & qdr   & rel   & skt   & AVG \\
			\midrule
			clp   & -     & 16.90  & 35.30  & 10.80  & 53.50  & 36.90  & 30.70  & clp   & -     & 18.15  & 35.75  & 6.83  & 51.70  & 40.42  & 30.57  \\
			inf   & 27.80  & -     & 28.20  & 4.40  & 48.20  & 22.50  & 26.20  & inf   & 29.60  & -     & 34.40  & 2.30  & 48.40  & 26.00  & 28.14  \\
			pnt   & 37.70  & 14.50  & -     & 4.60  & 59.40  & 33.50  & 30.00  & pnt   & 43.70  & 15.50  & -     & 2.40  & 58.40  & 37.80  & 31.56  \\
			qdr   & 14.90  & 1.50  & 6.20  & -     & 10.90  & 10.20  & 8.70  & qdr   & 19.10  & 3.20  & 8.90  & -     & 14.00  & 15.50  & 12.14  \\
			rel   & 49.40  & 20.80  & 47.20  & 4.80  & -     & 38.20  & 32.00  & rel   & 49.70  & 21.10  & 52.00  & 1.80  & -     & 37.80  & 32.48  \\
			skt   & 50.10  & 16.50  & 43.70  & 11.10  & 55.60  & -     & 35.40  & skt   & 54.10  & 17.70  & 43.20  & 9.60  & 52.00  & -     & 35.32  \\
			AVG   & 36.00  & 14.00  & 32.10  & 7.10  & 45.50  & 28.30  & 27.20  & AVG   & 39.24  & 15.13  & 34.85  & 4.59  & 44.90  & 31.50  & \color{red}{\textbf{28.37}} \\
			\bottomrule
		\end{tabular}%
	}
	\label{tab:domainnet}%
\end{table*}%

\begin{figure*}[!t]
\centering
\includegraphics[width=1\linewidth]{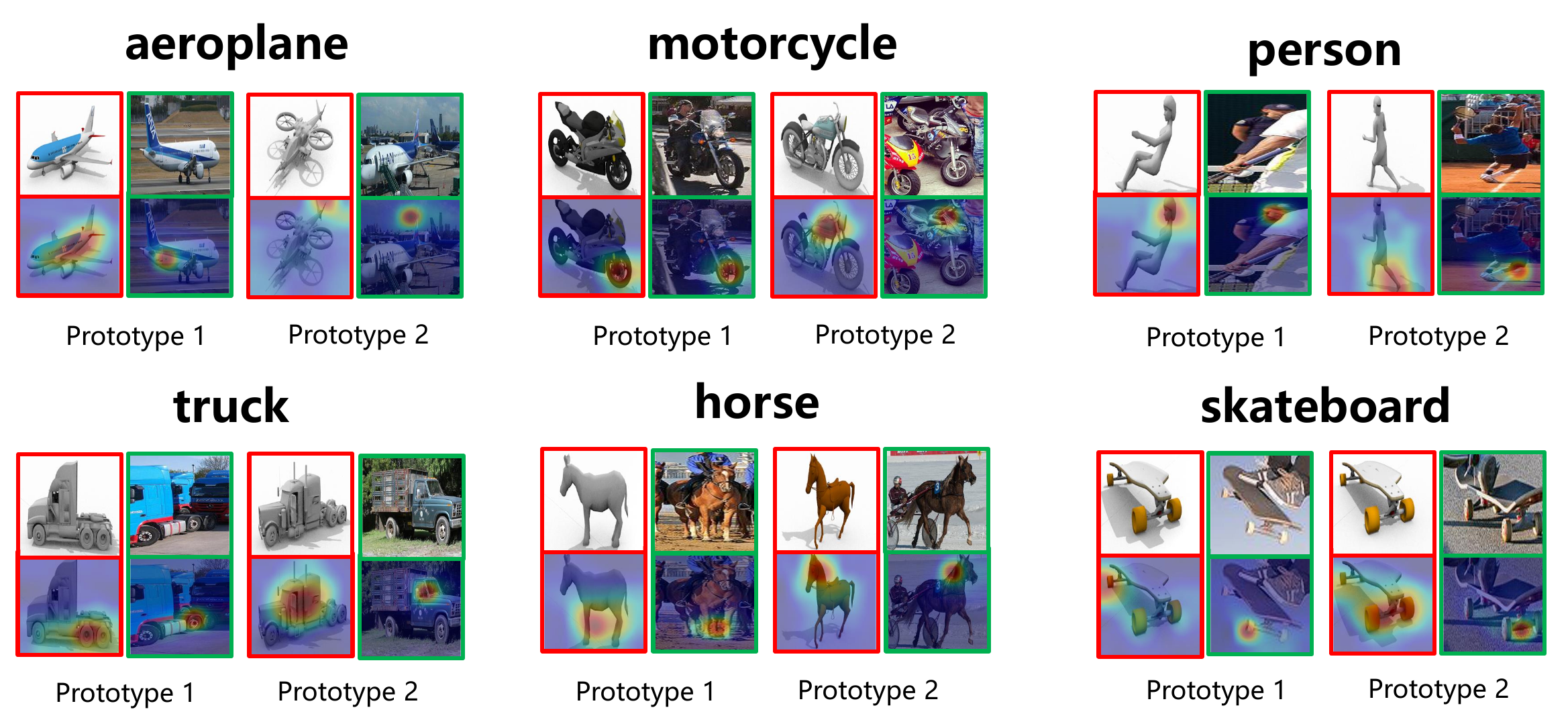}
\caption{The prototype visualizations in VisDA dataset. For each category, the top row represents the original images, and the bottom row highlights image patches related to prototypes.
Source and target domains are denoted by red and green boxes.}
\label{fig:pro}

\end{figure*}

\subsection{Recognition Results}


\noindent{\textbf{Results on Office-Home.}}
Table \ref{tab:officehome} shows the experimental results of 12 transfer tasks in the Office-Home dataset.
The best results are highlighted and the suboptimal results are underlined. 
Some standard self-training methods such as VAT and FixMatch perform well,
but they are easily affected by false label noise when suffering from a large domain gap
resulting in performance degradation. 
Although MDD, GSDA, and RSDA-MSTN have designed complex domain alignment strategies, their performance is still lower than ours. 
%
%
Compared with CST, our proposed TCPL outperforms it by an averaged absolute gain of $1.32\%$, which indicates that the interpretable prototype learning strategy and the self-predictive consistent pseudo-label strategy and enhance the model's transferability.


\begin{figure}[!t]
	\subfigure[]{
		\begin{minipage}[t]{1\linewidth}
			\label{fig:vis:visda}
			\centering
			\includegraphics[width=\linewidth]{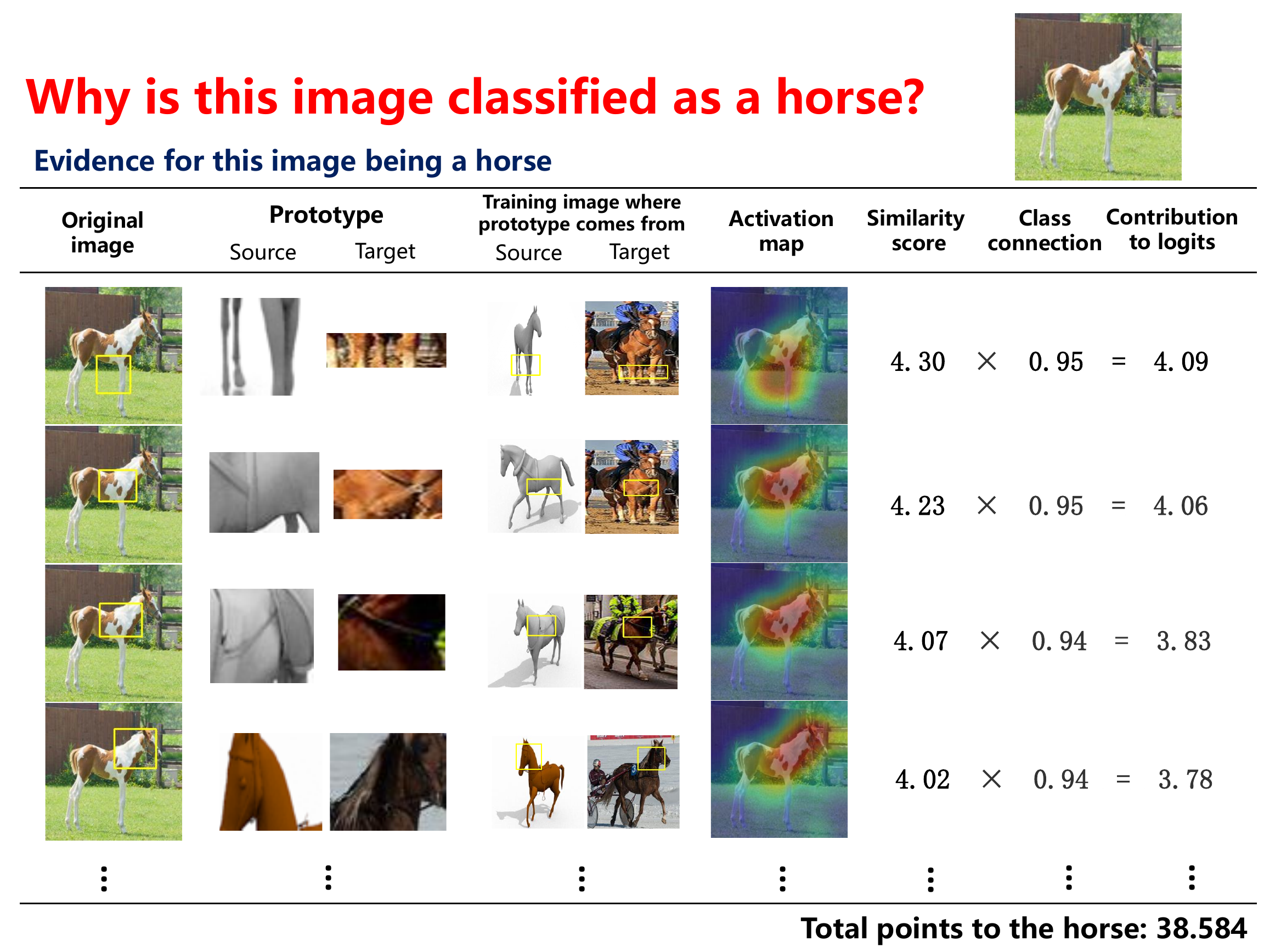}
		\end{minipage}
	}\\
	\subfigure[]{
		\begin{minipage}[t]{1\linewidth}
			\label{fig:vis:domainnet}
			\centering
			\includegraphics[width=\linewidth]{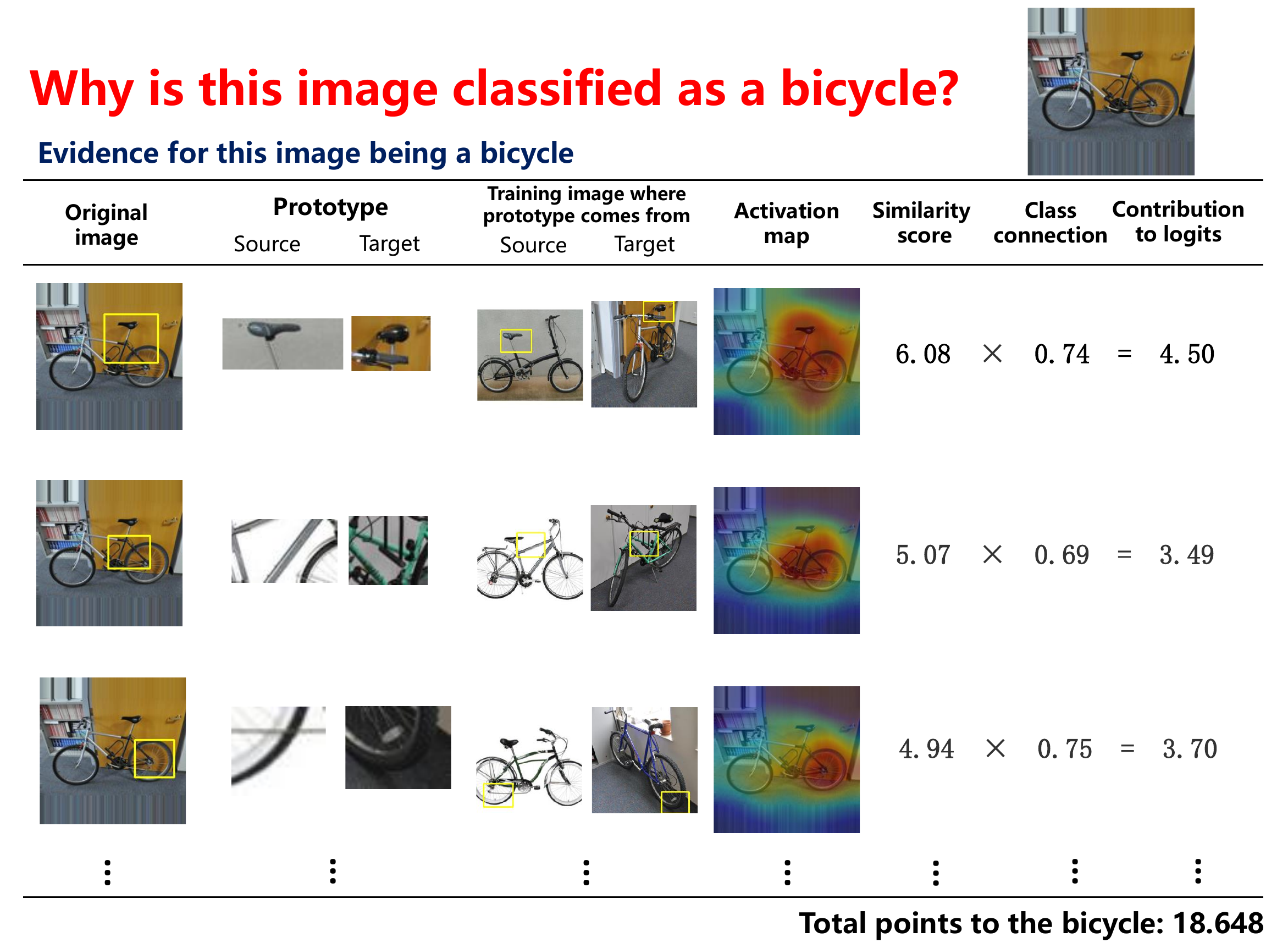}
		\end{minipage}
	}\\
	\subfigure[]{
		\begin{minipage}[t]{1\linewidth}
			\label{fig:tsne:officehome}
			\centering
			\includegraphics[width=\linewidth]{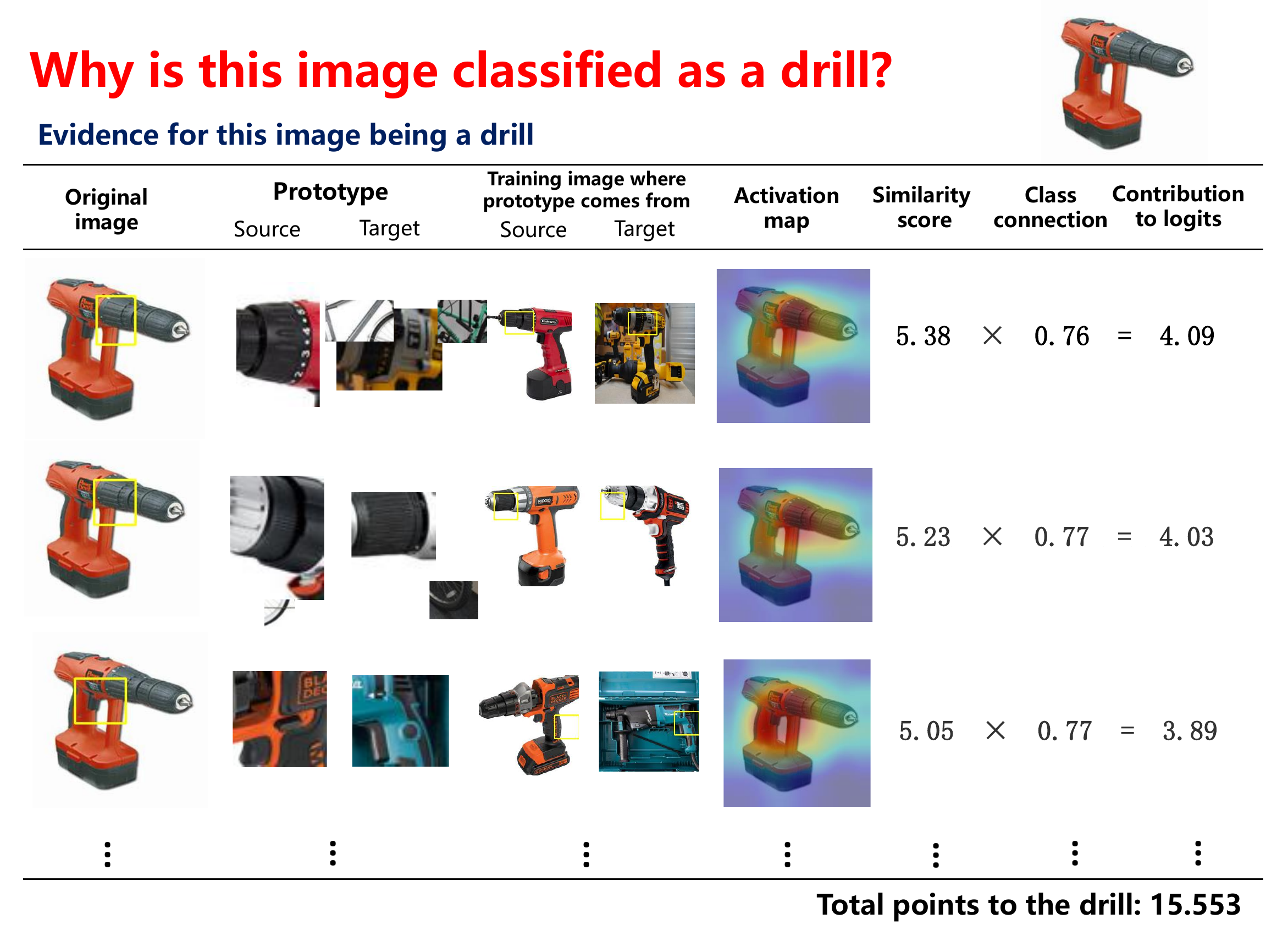}
		\end{minipage}
	}
	\caption{The interpretable reasoning process of identifying categories.}
	\label{fig:inference}
\end{figure}


\noindent{\textbf{Results on VisDA and DomainNet.}}
In Table \ref{tab:visda} and Table \ref{tab:domainnet}, we present the detailed results on two large-scale datasets, VisDA and DomainNet.
In VisDA, the model is required to associate a large domain gap, \ie, synthetic images and real images.
Our method surpasses all compared methods with the ResNet-50 and ResNet-101 backbones, especially, CST and SENTRAY.
Since the proposed self-predictive consistent pseudo-label strategy learns model with augmented samples, which can be regarded as a self-supervised learning strategy,
we also compare models combining domain adaptation and self-training methods such as DIRT-T, CDAN+VAT+Entropy, and MDD+FixMatch. 
Despite this, our method still perform the best, 
indicating that the proposed pseudo-label strategy is superior to simple combinations of self-training and domain adaptation strategies.
To involve more categories and larger domain gaps for evaluation, we examine our approach on the DomainNet dataset in Table \ref{tab:domainnet}. 
The average classification accuracy of our approach is 28\% among 30 transfer tasks and outperforms previous state-of-the-art method CGDM more than 1\%.
Since DomainNet is extremely challenging, 
the 1\% improvement demonstrates the superiority and robustness of our approach on large-scale dataset.

\begin{table*}[!t]
	\centering
	\caption{Classification accuracy (\%) of ablation studies in hierarchically prototypical module on Office-Home dataset.}
	\scalebox{1}{
		\begin{tabular}{cccccccccccccc}
			\toprule
			Source    & Ar    & Ar    & Ar    & Cl    & Cl    & Cl    & Pr    & Pr    & Pr    & Rw    & Rw    & Rw & \multirow{2}[0]{*}{AVG} \\
			Target   & Cl    & Pr    & Rw    & Ar    & Pr    & Rw    & Ar    & Cl    & Rw    & Ar    & Cl    & Pr    &  \\
			\midrule
			w/o HPM & 55.30  & 76.80  & 80.50  & 64.60  & 73.40  & 72.80  & 66.40  & 55.20  & 80.70  & 74.20  & 60.00  & 84.60  & $70.38_{\downarrow 0.61}$  \\
			Baseline & 56.90  & 77.50  & 80.60  & 65.00  & 74.30  & 73.40  & 67.40  & 55.30  & 81.10  & 74.80  & 60.90  & 84.70  & 70.99  \\
			\hline 
			w/o HPM & 60.30  & 80.00  & 82.30  & 67.80  & 73.90  & 75.40  & 71.40  & 59.70  & 83.50  & 77.60  & 64.80  & 86.30  & $73.58_{\downarrow 0.74}$  \\
			TCPL & 61.20  & 80.50  & 82.80  & 68.80  & 75.10  & 76.50  & 71.70  & 59.80  & 83.50  & 78.10  & 66.20  & 87.60  & 74.32  \\
			\bottomrule
		\end{tabular}%
	}
	\label{tab:ablation_spf}%
\end{table*}%

\begin{table*}[!t]
	\centering
	\caption{The classification accuracy (\%) of ablation studies in self-predictive consistent pseudo-label strategy on Office-Home dataset. CC, PD, and PT denote classification confidence, prediction, and prototype criteria, respectively. }
	\small
	\scalebox{0.9}{
		\renewcommand{\arraystretch}{1.1}
		\begin{tabular}{ccc|ccccccccccccc}
			\toprule
			\multirow{2}[0]{*}{CC} & \multirow{2}[0]{*}{PD} & \multirow{2}[0]{*}{PT} & Ar    & Ar    & Ar    & Cl    & Cl    & Cl    & Pr    & Pr    & Pr    & Rw    & Rw    & Rw    & \multirow{2}[0]{*}{AVG} \\
			&     &      & Cl    & Pr    & Rw    & Ar    & Pr    & Rw    & Ar    & Cl    & Rw    & Ar    & Cl    & Pr    &  \\
			\midrule
			\XSolidBrush   & \XSolidBrush   & \XSolidBrush       & 56.90  & 77.50  & 80.60  & 65.00  & 74.30  & 73.40  & 67.40  & 55.30  & 81.10  & 74.80  & 60.90  & 84.70  & 70.99 ($+0.00$)  \\
			\CheckmarkBold     & \XSolidBrush   & \XSolidBrush       & 59.00  & 78.10  & 82.50  & 66.30  & 74.80  & 73.70  & 71.70  & 58.50  & 82.70  & 77.40  & 62.60  & 85.90  & 72.77 ($+1.78$)\\
			\XSolidBrush   & \CheckmarkBold     & \XSolidBrush       & 56.80  & 77.80  & 81.20  & 62.40  & 71.60  & 67.60  & 66.50  & 55.60  & 81.00  & 75.60  & 60.30  & 84.00  & 70.03 ($-0.96$) \\
			\XSolidBrush   & \XSolidBrush   & \CheckmarkBold         & 50.30  & 68.70  & 75.60  & 55.50  & 63.90  & 62.50  & 62.30  & 51.50  & 77.70  & 74.00  & 58.40  & 80.90  & 65.11 ($-5.88$)\\
			\CheckmarkBold     & \CheckmarkBold     & \XSolidBrush       & 56.70  & 78.70  & 81.30  & 64.60  & 72.80  & 72.40  & 69.10  & 55.40  & 81.60  & 75.50  & 62.10  & 84.30  & 71.21 ($+0.22$)\\
			\CheckmarkBold     & \XSolidBrush   & \CheckmarkBold         & 57.80  & 76.70  & 79.30  & 65.60  & 73.50  & 73.10  & 67.00  & 56.50  & 81.00  & 76.10  & 61.30  & 83.30  & 70.93 ($-0.06$)\\
			\XSolidBrush   & \CheckmarkBold     & \CheckmarkBold         & 55.80  & 75.80  & 81.10  & 54.80  & 69.20  & 73.90  & 66.40  & 52.70  & 82.30  & 74.70  & 57.30  & 84.60  & 69.05 ($-1.94$)\\
			\CheckmarkBold     & \CheckmarkBold     & \CheckmarkBold         & 61.20  & 80.50  & 82.80  & 68.80  & 75.10  & 76.50  & 71.70  & 59.80  & 83.50  & 78.10  & 66.20  & 87.60  & 74.32 ($+3.33$)\\
			\midrule
		\end{tabular}%
	}
	\label{tab:ablation_}%

\end{table*}%

\subsection{Interpretability Analysis}

\noindent{\textbf{Prototype Visualizations.}}
To verify whether the proposed method could learn conceptual prototypes
and transfer categorical basic concepts from the source domain to the target domain,
we propose to visualize the learned prototypes with source and target training images.
Given a trained TCPL model,
the learned prototypes are already projected to the closest latent patch of some training images.
Therefore, a prototype $\mathbf{p}_j$ is exactly equal to some patches of the latent representations $G_f(\textbf{x})$ of source and target training images $\textbf{x} \in \mathcal{D}_S \cup \mathcal{D}_T$. 
Since the patch of $\mathbf{x}$ corresponding to the prototype $\mathbf{p}_j$ should be the region where the prototype 
$\mathbf{p}_j$ most strongly activates, 
we visualize the prototype $\mathbf{p}_j$ by first obtaining the activation map of $\mathbf{x}$ by the prototype $\mathbf{p}_j$: 
this can be done by forwarding $\mathbf{x}$ through the trained network
and upsampling the activation map produced by the prototype unit $G_{\mathbf{p}_j}$ to the size of the image $\mathbf{x}$. 
After we obtain such an activation map, we can locate the patch of  $\mathbf{x}$ on which $\mathbf{p}_j$ has the strongest activation by finding the high activation region in the (upsampled) activation map.
In our experiments, we define the high activation region in an upsampled activation map as the smallest rectangular region that encloses pixels whose corresponding activation value in the aforementioned activation map is at least 
95\% of all activation values in that same map. 
Finally, we can visualize the prototype  $G_{\mathbf{p}_j}$ using the image patch  that corresponds to the high activation region.

We visualize the learned prototypes respectively with source and target images in the VisDA dataset as shown in Figure \ref{fig:pro}, 
and obtain some interesting observations:  
(1) The learned prototypes do encode the information of categorical basic concepts,
such as the fuselage and tail of ``aeroplane'', 
the wheel and seat of ``motorcycle'', the head and leg of ``person'', 
the wheel and window of ``truck'', the leg and head of ``horse'', the wheel and axle of ``skateboard'', etc.
(2) The conceptual meanings of the same prototype are consistent in different domains,
for example, prototype 1 of ``aeroplane'' both represents ``fuselage'' in source and target domains.
It reveals that the proposed method could transfer categorical basic concepts to the target domain and assist to conduct interpretable classification in the target domain.

\noindent{\textbf{Interpretable Inference.}}
%
Figure \ref{fig:inference} shows the interpretable inference process of the proposed method on test images of several classes, \ie, horse (from VisDA dataset), bicycle (from DomainNet dataset), and drill (from Office-Home dataset).
Given a test image $\textbf{x}$, our model tries to find evidence for  $\textbf{x}$ to be of class
by comparing its feature maps with every learned prototype.
We take the Figure \ref{fig:vis:visda} as an example to illustrate the details of how and why this image is classified as a horse.
Specifically, our model tries to find evidence for the horse class by comparing the image's latent patches with each prototype (visualized in the ``Prototype'' column) of that class.
As shown in the ``Activation map'' column,
the first prototype of the horse class activates most strongly on the leg of the testing image,
and the second prototype on the belly, etc.
The most activated image patch of the given image for each prototype is marked by a bounding box in the ``Original image'' column and this is the image patch that the model considers to look like the corresponding prototype.
In this case, our model finds a high similarity between the leg of the given image and the prototypical leg of a horse (with a similarity score of 4.09), 
as well as between the belly and the prototypical belly (with a similarity score of 4.06). 
These similarity scores are weighted and summed together to give a final score for the image belonging to the horse class. 
The reasoning process is similar for all other classes.

\subsection{Ablation Studies}

\noindent{\textbf{Effect of Hierarchically Prototypical Module\footnote{Note that this module indicates the multi-scale feature learning process in Section~\ref{sec:hpm}.}.}}
To testify whether the hierarchically prototypical module (HPM) could encode effective multi-scale information when learning conceptual prototypes, 
we compare the performance of two models (baseline and TCPL) with and without HPM. The baseline model is obtained by removing the self-predictive consistent pseudo-label strategy from the full TCPL method.
The experimental results on the Office-Home dataset are shown in Table \ref{tab:ablation_spf}. 
Without hierarchical features, the average classification accuracy of TCPL and baseline method is reduced by 0.74\% and 0.61\% respectively,
indicating that the multi-scale information captured by HPM facilitates better target representations. Here, ``w/o HPM'' means that we do not adopt the multi-scale feature extraction operation for input images, while prototypes are stilled used for learning interpretable domain adaptation.

\noindent{\textbf{Effect of Prototype Learning.}}
To validate the effectiveness of our proposed interpretable prototype learning strategy, we design a baseline variant, namely w/o PL (Prototype Learning), which abandon the prototype learning loss ${\mathcal L}_{\text{cdpd}}$ and ${\mathcal L}_{\text{dd}}$. In this baseline, prototypes are unlearnable and only updated by using the proposed prototype transparency strategy. Since $\mathbf{W}$ in this baseline is meaningless, to perform image classification, we employ the classifier in CST~\cite{liu2021cycle} for implementing $G_c$. From~Table \ref{tab:pl} we can observe that the prototype learning strategy is useful in our proposed framework. Compared with CST, we can also conclude that the self-predictive
consistent pseudo label mining strategy is effective.

\begin{table*}[htbp]
	\centering
	\caption{Classification Accuracy (\%) of ablation studies about the prototype learning strategy on Office-Home dataset.}
	\small
	\scalebox{1}{
		\renewcommand{\arraystretch}{1.1}
		\begin{tabular}{cccccccccccccc}
			\toprule
			Source & Ar    & Ar    & Ar    & Cl    & Cl    & Cl    & Pr    & Pr    & Pr    & Rw    & Rw    & Rw    & \multirow{2}[0]{*}{AVG} \\
			Target & Cl    & Pr    & Rw    & Ar    & Pr    & Rw    & Ar    & Cl    & Rw    & Ar    & Cl    & Pr    &  \\
			\midrule
			CST   & 59.00 & 79.60 & 83.40 & 68.40 & 77.10 & 76.70 & 68.90 & 56.40 & 83.00 & 75.30 & 62.20 & 85.10 & 73.00  \\ 
			w/o PL   & 60.90 & 78.90 & 82.40 & 68.30 & 73.90 & 76.10 & 70.40 & 59.30 & 83.10 & 77.20 & 64.90 & 85.90 & 73.44   \\ \midrule
			TCPL  & 61.20 & 80.50 & 82.80 & 68.80 & 75.10 & 76.50 & 71.70 & 59.80 & 83.50 & 78.10 & 66.20 & 87.60 & 74.32  \\
			\bottomrule
		\end{tabular}%
	}
	\label{tab:pl}%
	\vspace{-4mm}
\end{table*}%

\noindent{\textbf{Effect of Self-predictive Consistent Pseudo Label Strategy.}}
The proposed self-predictive consistent pseudo-label strategy designs three criteria to evaluate the reliability of samples' pseudo-labels,
which plays an important role in reducing the domain gap and learning domain-shared conceptual prototypes. 
To analyze the performance impact of different criteria, we design several variants of the proposed pseudo-label strategy 
and report their classification accuracy on the Office-Home dataset in Table \ref{tab:ablation_}.
According to the results in Table \ref{tab:ablation_},
we could observe that:
(1) The classification confidence is the most important criterion for the pseudo-label strategy due to its discriminative ability.
(2) Comparing the experimental results of methods using two criteria, 
the accuracy of methods combining classification confidence criterion and other criteria is almost the same as the reference accuracy, indicating that the confidence criterion cannot completely eliminate noisy annotations caused by prediction and prototype criteria. 
(3) When applying all the three criteria simultaneously, 
the performance gain is the largest, almost 3.33\%. 
This strategy can be understood as utilizing prediction and prototype criteria to filter out the wrongly labeled instances among the candidates sampled by the classification confidence criterion.

\begin{figure}[!t]
	\subfigure[Sensitivity analysis of $M$]{
		\begin{minipage}[t]{0.9\linewidth}
			\label{fig:m}
			\centering
			\includegraphics[width=\linewidth]{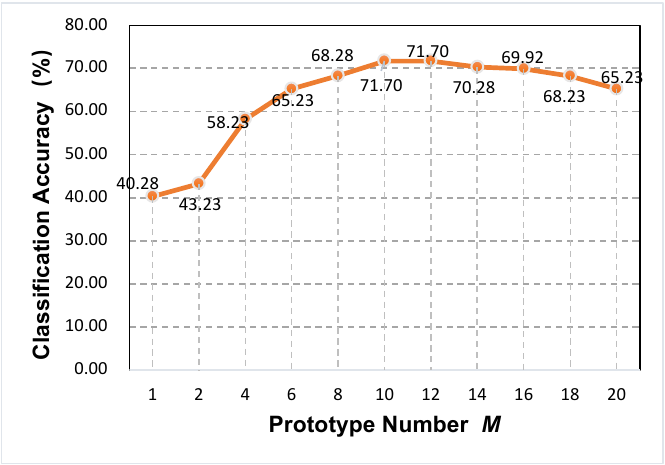}
		\end{minipage}
	}
	\subfigure[Sensitivity analysis of $V$]{
		\begin{minipage}[t]{0.9\linewidth}
			\label{fig:v}
			\centering
			\includegraphics[width=\linewidth]{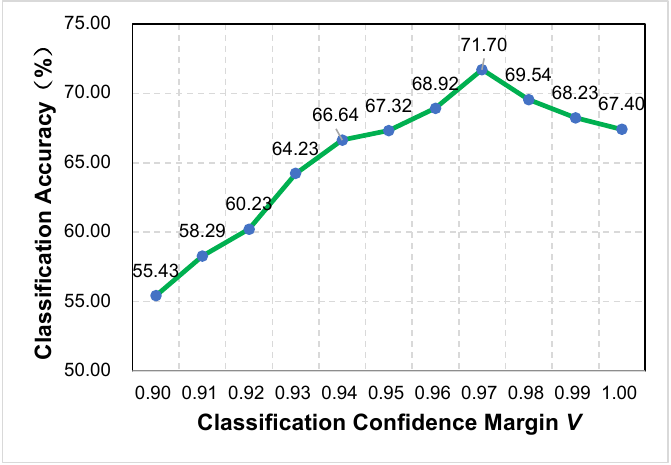}
		\end{minipage}
	}
	\caption{Effect of prototype number for each category $M$ and classification confidence margin $V$.}
	\label{fig:hyparam}
\end{figure}

\noindent{\textbf{Prototype number $M$ for each category.}}
The prototype number $M$ for each category determines the model capacity to encode categorical semantic concepts into prototypes,
and we analyze its effect on the Rr$\to$Ar task in the Office-home dataset.
As shown in Figure \ref{fig:m}, we could observe that fewer or more prototypes lead to lower classification accuracy.
Because a small number of prototypes cannot comprehensively learn the basic conceptual information for each category,
while too many prototypes are redundant, resulting in the overfitting issue of the source domain and damaging model performance in the target domain.
%

\noindent{\textbf{Classification confidence margin $V$.}}
The classification confidence margin $V$ in the proposed self-predictive consistent pseudo-label strategy aims to  
select target samples with confident pseudo labels for reducing the domain gap.
We analyze the effect of $V$ on the Rr$\to$Ar task in the Office-home dataset.
From Figure \ref{fig:v} we could find that the model performance significantly drops when the value of $V$ is smaller than 0.95. 
The reason is that a smaller $V$ may introduces more pseudo-label noise thus harms the effectiveness of cross-domain learning. In addition, a too large threshold can hardly select enough target samples for joint training, which also makes the performance inferior. 
%

\section{Conclusion}

In this paper, we explore an inherently interpretable deep method, named Transferable Conceptual Prototype Learning (TCPL), for unsupervised domain adaptation.
To learn transferable conceptual prototypes,
we propose a hierarchically prototypical module learned with an
interpretable prototype learning strategy and a self-predictive
consistent pseudo-label strategy, making sure explainable decision
knowledge from the source domain could be reliably transferred to
the unlabeled target domain.
Comprehensive experiments provide reliable interpretations and prove the effectiveness of the proposed method.
Several limitations of this paper are noteworthy. 
In this paper, we adopt a fixed number of mutually exclusive prototypes for each class. Although simple and effective, learning category-shared and specific prototypes adaptively can be further considered to improve the performance and interpretability. Besides, exploring the relations between prototypes can also facilitate the performance improvement and the extension of our framework to DA-based multi-label classification problem.

{\small
	\bibliographystyle{IEEEtran}
	\bibliography{egbib}

\begin{thebibliography}{10}
\providecommand{\url}[1]{#1}
\csname url@samestyle\endcsname
\providecommand{\newblock}{\relax}
\providecommand{\bibinfo}[2]{#2}
\providecommand{\BIBentrySTDinterwordspacing}{\spaceskip=0pt\relax}
\providecommand{\BIBentryALTinterwordstretchfactor}{4}
\providecommand{\BIBentryALTinterwordspacing}{\spaceskip=\fontdimen2\font plus
\BIBentryALTinterwordstretchfactor\fontdimen3\font minus
  \fontdimen4\font\relax}
\providecommand{\BIBforeignlanguage}[2]{{%
\expandafter\ifx\csname l@#1\endcsname\relax
\typeout{** WARNING: IEEEtran.bst: No hyphenation pattern has been}%
\typeout{** loaded for the language `#1'. Using the pattern for}%
\typeout{** the default language instead.}%
\else
\language=\csname l@#1\endcsname
\fi
#2}}
\providecommand{\BIBdecl}{\relax}
\BIBdecl

\bibitem{cai2019learning}
R.~Cai, Z.~Li, P.~Wei, J.~Qiao, K.~Zhang, and Z.~Hao, ``Learning disentangled
  semantic representation for domain adaptation,'' in \emph{IJCAI: proceedings
  of the conference}, vol. 2019, 2019, p. 2060.

\bibitem{cui2020towards}
S.~Cui, S.~Wang, J.~Zhuo, L.~Li, Q.~Huang, and Q.~Tian, ``Towards
  discriminability and diversity: Batch nuclear-norm maximization under label
  insufficient situations,'' in \emph{Proceedings of the IEEE/CVF Conference on
  Computer Vision and Pattern Recognition}, 2020, pp. 3941--3950.

\bibitem{liu2021cycle}
H.~Liu, J.~Wang, and M.~Long, ``Cycle self-training for domain adaptation,''
  \emph{Advances in Neural Information Processing Systems}, vol.~34, 2021.

\bibitem{chen2022reusing}
L.~Chen, H.~Chen, Z.~Wei, X.~Jin, X.~Tan, Y.~Jin, and E.~Chen, ``Reusing the
  task-specific classifier as a discriminator: Discriminator-free adversarial
  domain adaptation,'' in \emph{Proceedings of the IEEE/CVF Conference on
  Computer Vision and Pattern Recognition}, 2022, pp. 7181--7190.

\bibitem{rangwani2022closer}
H.~Rangwani, S.~K. Aithal, M.~Mishra, A.~Jain, and V.~B. Radhakrishnan, ``A
  closer look at smoothness in domain adversarial training,'' in
  \emph{International Conference on Machine Learning}.\hskip 1em plus 0.5em
  minus 0.4em\relax PMLR, 2022, pp. 18\,378--18\,399.

\bibitem{zeiler2014visualizing}
M.~D. Zeiler and R.~Fergus, ``Visualizing and understanding convolutional
  networks,'' in \emph{European conference on computer vision}, 2014, pp.
  818--833.

\bibitem{zintgraf2017visualizing}
L.~M. Zintgraf, T.~S. Cohen, T.~Adel, and M.~Welling, ``Visualizing deep neural
  network decisions: Prediction difference analysis,'' \emph{arXiv:1702.04595},
  2017.

\bibitem{zhou2016learning}
B.~Zhou, A.~Khosla, A.~Lapedriza, A.~Oliva, and A.~Torralba, ``Learning deep
  features for discriminative localization,'' in \emph{Proceedings of the IEEE
  conference on computer vision and pattern recognition}, 2016, pp. 2921--2929.

\bibitem{gao2017deep}
J.~Gao, T.~Zhang, X.~Yang, and C.~Xu, ``Deep relative tracking,'' \emph{IEEE
  Transactions on Image Processing}, vol.~26, no.~4, pp. 1845--1858, 2017.

\bibitem{gilpin2018explaining}
L.~H. Gilpin, D.~Bau, B.~Z. Yuan, A.~Bajwa, M.~Specter, and L.~Kagal,
  ``Explaining explanations: An overview of interpretability of machine
  learning,'' in \emph{Proceedings of the IEEE International Conference on data
  science and advanced analytics}, 2018, pp. 80--89.

\bibitem{selvaraju2017grad}
R.~R. Selvaraju, M.~Cogswell, A.~Das, R.~Vedantam, D.~Parikh, and D.~Batra,
  ``Grad-cam: Visual explanations from deep networks via gradient-based
  localization,'' in \emph{Proceedings of the IEEE international conference on
  computer vision}, 2017, pp. 618--626.

\bibitem{ribeiro2016model}
M.~T. Ribeiro, S.~Singh, and C.~Guestrin, ``Model-agnostic interpretability of
  machine learning,'' \emph{arXiv preprint arXiv:1606.05386}, 2016.

\bibitem{wang2020score}
H.~Wang, Z.~Wang, M.~Du, F.~Yang, Z.~Zhang, S.~Ding, P.~Mardziel, and X.~Hu,
  ``Score-cam: Score-weighted visual explanations for convolutional neural
  networks,'' in \emph{Proceedings of the IEEE/CVF conference on computer
  vision and pattern recognition workshops}, 2020, pp. 24--25.

\bibitem{chen2019looks}
C.~Chen, O.~Li, D.~Tao, A.~Barnett, C.~Rudin, and J.~K. Su, ``This looks like
  that: deep learning for interpretable image recognition,'' \emph{Advances in
  neural information processing systems}, vol.~32, 2019.

\bibitem{ge2021peek}
Y.~Ge, Y.~Xiao, Z.~Xu, M.~Zheng, S.~Karanam, T.~Chen, L.~Itti, and Z.~Wu, ``A
  peek into the reasoning of neural networks: Interpreting with structural
  visual concepts,'' in \emph{Proceedings of the IEEE/CVF Conference on
  Computer Vision and Pattern Recognition}, 2021, pp. 2195--2204.

\bibitem{wang2021interpretable}
J.~Wang, H.~Liu, X.~Wang, and L.~Jing, ``Interpretable image recognition by
  constructing transparent embedding space,'' in \emph{Proceedings of the
  IEEE/CVF International Conference on Computer Vision}, 2021, pp. 895--904.

\bibitem{liang2020training}
H.~Liang, Z.~Ouyang, Y.~Zeng, H.~Su, Z.~He, S.-T. Xia, J.~Zhu, and B.~Zhang,
  ``Training interpretable convolutional neural networks by differentiating
  class-specific filters,'' in \emph{European Conference on Computer Vision},
  2020, pp. 622--638.

\bibitem{hou2021visualizing}
Y.~Hou and L.~Zheng, ``Visualizing adapted knowledge in domain transfer,'' in
  \emph{Proceedings of the IEEE/CVF Conference on Computer Vision and Pattern
  Recognition}, 2021, pp. 13\,824--13\,833.

\bibitem{zunino2021explainable}
A.~Zunino, S.~A. Bargal, R.~Volpi, M.~Sameki, J.~Zhang, S.~Sclaroff, V.~Murino,
  and K.~Saenko, ``Explainable deep classification models for domain
  generalization,'' in \emph{Proceedings of the IEEE/CVF Conference on Computer
  Vision and Pattern Recognition}, 2021, pp. 3233--3242.

\bibitem{petryk2022guiding}
S.~Petryk, L.~Dunlap, K.~Nasseri, J.~Gonzalez, T.~Darrell, and A.~Rohrbach,
  ``On guiding visual attention with language specification,'' in
  \emph{Proceedings of the IEEE/CVF Conference on Computer Vision and Pattern
  Recognition}, 2022, pp. 18\,092--18\,102.

\bibitem{tenenbaum2011grow}
J.~B. Tenenbaum, C.~Kemp, T.~L. Griffiths, and N.~D. Goodman, ``How to grow a
  mind: Statistics, structure, and abstraction,'' \emph{Science}, vol. 331, no.
  6022, pp. 1279--1285, 2011.

\bibitem{stach2004local}
S.~Stach, J.~Benard, and M.~Giurfa, ``Local-feature assembling in visual
  pattern recognition and generalization in honeybees,'' \emph{Nature}, vol.
  429, no. 6993, pp. 758--761, 2004.

\bibitem{long2015learning}
M.~Long, Y.~Cao, J.~Wang, and M.~Jordan, ``Learning transferable features with
  deep adaptation networks,'' in \emph{International Conference on Machine
  Learning}, 2015, pp. 97--105.

\bibitem{long2017deep}
M.~Long, H.~Zhu, J.~Wang, and M.~I. Jordan, ``Deep transfer learning with joint
  adaptation networks,'' in \emph{International Conference on Machine
  Learning}, 2017, pp. 2208--2217.

\bibitem{sun2016deep}
B.~Sun and K.~Saenko, ``Deep coral: Correlation alignment for deep domain
  adaptation,'' in \emph{European conference on computer vision}, 2016, pp.
  443--450.

\bibitem{li2021implicit}
M.~Li, K.~Jiang, and X.~Zhang, ``Implicit task-driven probability discrepancy
  measure for unsupervised domain adaptation,'' \emph{Advances in neural
  information processing systems}, vol.~34, pp. 25\,824--25\,838, 2021.

\bibitem{hu2022learning}
J.~Hu, H.~Zhong, F.~Yang, S.~Gong, G.~Wu, and J.~Yan, ``Learning unbiased
  transferability for domain adaptation by uncertainty modeling,'' in
  \emph{Computer Vision--ECCV 2022: 17th European Conference, Tel Aviv, Israel,
  October 23--27, 2022, Proceedings, Part XXXI}.\hskip 1em plus 0.5em minus
  0.4em\relax Springer, 2022, pp. 223--241.

\bibitem{wang2022probability}
W.~Wang, Z.~Shen, D.~Li, P.~Zhong, and Y.~Chen, ``Probability-based graph
  embedding cross-domain and class discriminative feature learning for domain
  adaptation,'' \emph{IEEE Transactions on Image Processing}, vol.~32, pp.
  72--87, 2022.

\bibitem{tzeng2014deep}
E.~Tzeng, J.~Hoffman, N.~Zhang, K.~Saenko, and T.~Darrell, ``Deep domain
  confusion: Maximizing for domain invariance,'' \emph{arXiv preprint
  arXiv:1412.3474}, 2014.

\bibitem{goodfellow2014generative}
I.~Goodfellow, J.~Pouget-Abadie, M.~Mirza, B.~Xu, D.~Warde-Farley, S.~Ozair,
  A.~Courville, and Y.~Bengio, ``Generative adversarial nets,'' in
  \emph{Advances in neural information processing systems}, 2014, pp.
  2672--2680.

\bibitem{ganin2015unsupervised}
Y.~Ganin and V.~Lempitsky, ``Unsupervised domain adaptation by
  backpropagation,'' in \emph{International Conference on Machine Learning},
  2015, pp. 1180--1189.

\bibitem{li2019joint}
S.~Li, C.~H. Liu, B.~Xie, L.~Su, Z.~Ding, and G.~Huang, ``Joint adversarial
  domain adaptation,'' in \emph{Proceedings of the 27th ACM International
  Conference on Multimedia}, 2019, pp. 729--737.

\bibitem{long2018conditional}
M.~Long, Z.~Cao, J.~Wang, and M.~I. Jordan, ``Conditional adversarial domain
  adaptation,'' \emph{Advances in neural information processing systems},
  vol.~31, 2018.

\bibitem{dai2021disentangling}
P.~Dai, P.~Chen, Q.~Wu, X.~Hong, Q.~Ye, Q.~Tian, C.-W. Lin, and R.~Ji,
  ``Disentangling task-oriented representations for unsupervised domain
  adaptation,'' \emph{IEEE Transactions on Image Processing}, vol.~31, pp.
  1012--1026, 2021.

\bibitem{tang2020unsupervised}
H.~Tang, K.~Chen, and K.~Jia, ``Unsupervised domain adaptation via structurally
  regularized deep clustering,'' in \emph{Proceedings of the IEEE/CVF
  conference on computer vision and pattern recognition}, 2020, pp. 8725--8735.

\bibitem{li2020enhanced}
M.~Li, Y.-M. Zhai, Y.-W. Luo, P.-F. Ge, and C.-X. Ren, ``Enhanced transport
  distance for unsupervised domain adaptation,'' in \emph{Proceedings of the
  IEEE/CVF Conference on Computer Vision and Pattern Recognition}, 2020, pp.
  13\,936--13\,944.

\bibitem{li2020domain}
S.~Li, C.~Liu, Q.~Lin, B.~Xie, Z.~Ding, G.~Huang, and J.~Tang, ``Domain
  conditioned adaptation network,'' in \emph{Proceedings of the AAAI Conference
  on Artificial Intelligence}, vol.~34, no.~07, 2020, pp. 11\,386--11\,393.

\bibitem{pan2019transferrable}
Y.~Pan, T.~Yao, Y.~Li, Y.~Wang, C.-W. Ngo, and T.~Mei, ``Transferrable
  prototypical networks for unsupervised domain adaptation,'' in
  \emph{Proceedings of the IEEE/CVF Conference on Computer Vision and Pattern
  Recognition}, 2019, pp. 2239--2247.

\bibitem{deng2021joint}
W.~Deng, Q.~Liao, L.~Zhao, D.~Guo, G.~Kuang, D.~Hu, and L.~Liu, ``Joint
  clustering and discriminative feature alignment for unsupervised domain
  adaptation,'' \emph{IEEE Transactions on Image Processing}, vol.~30, pp.
  7842--7855, 2021.

\bibitem{zhou2015predicting}
J.~Zhou and O.~G. Troyanskaya, ``Predicting effects of noncoding variants with
  deep learning--based sequence model,'' \emph{Nature methods}, vol.~12,
  no.~10, pp. 931--934, 2015.

\bibitem{petsiuk2018rise}
V.~Petsiuk, A.~Das, and K.~Saenko, ``Rise: Randomized input sampling for
  explanation of black-box models,'' \emph{arXiv preprint arXiv:1806.07421},
  2018.

\bibitem{ribeiro2016should}
M.~T. Ribeiro, S.~Singh, and C.~Guestrin, ``" why should i trust you?"
  explaining the predictions of any classifier,'' in \emph{Proceedings of the
  22nd ACM SIGKDD international conference on knowledge discovery and data
  mining}, 2016, pp. 1135--1144.

\bibitem{chattopadhay2018grad}
A.~Chattopadhay, A.~Sarkar, P.~Howlader, and V.~N. Balasubramanian,
  ``Grad-cam++: Generalized gradient-based visual explanations for deep
  convolutional networks,'' in \emph{2018 IEEE winter conference on
  applications of computer vision (WACV)}.\hskip 1em plus 0.5em minus
  0.4em\relax IEEE, 2018, pp. 839--847.

\bibitem{fukui2019attention}
H.~Fukui, T.~Hirakawa, T.~Yamashita, and H.~Fujiyoshi, ``Attention branch
  network: Learning of attention mechanism for visual explanation,'' in
  \emph{Proceedings of the IEEE/CVF Conference on Computer Vision and Pattern
  Recognition}, 2019, pp. 10\,705--10\,714.

\bibitem{li2018tell}
K.~Li, Z.~Wu, K.-C. Peng, J.~Ernst, and Y.~Fu, ``Tell me where to look: Guided
  attention inference network,'' in \emph{Proceedings of the IEEE Conference on
  Computer Vision and Pattern Recognition}, 2018, pp. 9215--9223.

\bibitem{zheng2019re}
M.~Zheng, S.~Karanam, Z.~Wu, and R.~J. Radke, ``Re-identification with
  consistent attentive siamese networks,'' in \emph{Proceedings of the IEEE/CVF
  Conference on Computer Vision and Pattern Recognition}, 2019, pp. 5735--5744.

\bibitem{wang2019sharpen}
L.~Wang, Z.~Wu, S.~Karanam, K.-C. Peng, R.~V. Singh, B.~Liu, and D.~N. Metaxas,
  ``Sharpen focus: Learning with attention separability and consistency,'' in
  \emph{Proceedings of the IEEE/CVF International Conference on Computer
  Vision}, 2019, pp. 512--521.

\bibitem{ghorbani2019towards}
A.~Ghorbani, J.~Wexler, J.~Y. Zou, and B.~Kim, ``Towards automatic
  concept-based explanations,'' \emph{Advances in Neural Information Processing
  Systems}, vol.~32, pp. 9277--9286, 2019.

\bibitem{bau2017network}
D.~Bau, B.~Zhou, A.~Khosla, A.~Oliva, and A.~Torralba, ``Network dissection:
  Quantifying interpretability of deep visual representations,'' in
  \emph{Proceedings of the IEEE conference on computer vision and pattern
  recognition}, 2017, pp. 6541--6549.

\bibitem{gonzalez2018semantic}
A.~Gonzalez-Garcia, D.~Modolo, and V.~Ferrari, ``Do semantic parts emerge in
  convolutional neural networks?'' \emph{International Journal of Computer
  Vision}, vol. 126, no.~5, pp. 476--494, 2018.

\bibitem{zhang2018interpretable}
Q.~Zhang, Y.~N. Wu, and S.-C. Zhu, ``Interpretable convolutional neural
  networks,'' in \emph{Proceedings of the IEEE conference on computer vision
  and pattern recognition}, 2018, pp. 8827--8836.

\bibitem{zhang2022explaining}
Y.~Zhang, T.~Yao, Z.~Qiu, and T.~Mei, ``Explaining cross-domain recognition
  with interpretable deep classifier,'' \emph{arXiv preprint arXiv:2211.08249},
  2022.

\bibitem{radford2021learning}
A.~Radford, J.~W. Kim, C.~Hallacy, A.~Ramesh, G.~Goh, S.~Agarwal, G.~Sastry,
  A.~Askell, P.~Mishkin, J.~Clark \emph{et~al.}, ``Learning transferable visual
  models from natural language supervision,'' in \emph{International conference
  on machine learning}.\hskip 1em plus 0.5em minus 0.4em\relax PMLR, 2021, pp.
  8748--8763.

\bibitem{shimodaira2000improving}
H.~Shimodaira, ``Improving predictive inference under covariate shift by
  weighting the log-likelihood function,'' \emph{Journal of statistical
  planning and inference}, vol.~90, no.~2, pp. 227--244, 2000.

\bibitem{ILSVRC15}
O.~Russakovsky, J.~Deng, H.~Su, J.~Krause, S.~Satheesh, S.~Ma, Z.~Huang,
  A.~Karpathy, A.~Khosla, M.~Bernstein, A.~C. Berg, and L.~Fei-Fei, ``{ImageNet
  Large Scale Visual Recognition Challenge},'' \emph{International Journal of
  Computer Vision}, vol. 115, no.~3, pp. 211--252, 2015.

\bibitem{he2016deep}
K.~He, X.~Zhang, S.~Ren, and J.~Sun, ``Deep residual learning for image
  recognition,'' in \emph{Proceedings of the IEEE/CVF Conference on Computer
  Vision and Pattern Recognition}, 2016, pp. 770--778.

\bibitem{girshick2015fast}
R.~Girshick, ``Fast r-cnn,'' in \emph{Proceedings of the IEEE international
  conference on computer vision}, 2015, pp. 1440--1448.

\bibitem{manders2018simple}
J.~Manders, E.~Marchiori, and T.~van Laarhoven, ``Simple domain adaptation with
  class prediction uncertainty alignment,'' \emph{arXiv:1804.04448}, 2018.

\bibitem{zhang2018collaborative}
W.~Zhang, W.~Ouyang, W.~Li, and D.~Xu, ``Collaborative and adversarial network
  for unsupervised domain adaptation,'' in \emph{Proceedings of the IEEE
  conference on computer vision and pattern recognition}, 2018, pp. 3801--3809.

\bibitem{kang2019contrastive}
G.~Kang, L.~Jiang, Y.~Yang, and A.~G. Hauptmann, ``Contrastive adaptation
  network for unsupervised domain adaptation,'' in \emph{Proceedings of the
  IEEE/CVF Conference on Computer Vision and Pattern Recognition}, 2019, pp.
  4893--4902.

\bibitem{tan2020class}
S.~Tan, X.~Peng, and K.~Saenko, ``Class-imbalanced domain adaptation: an
  empirical odyssey,'' in \emph{European Conference on Computer Vision}, 2020,
  pp. 585--602.

\bibitem{gao2019smart}
J.~Gao, T.~Zhang, and C.~Xu, ``Smart: Joint sampling and regression for visual
  tracking,'' \emph{IEEE Transactions on Image Processing}, vol.~28, no.~8, pp.
  3923--3935, 2019.

\bibitem{venkateswara2017deep}
H.~Venkateswara, J.~Eusebio, S.~Chakraborty, and S.~Panchanathan, ``Deep
  hashing network for unsupervised domain adaptation,'' in \emph{Proceedings of
  the IEEE/CVF Conference on Computer Vision and Pattern Recognition}, 2017,
  pp. 5018--5027.

\bibitem{8575439}
X.~{Peng}, B.~{Usman}, N.~{Kaushik}, D.~{Wang}, J.~{Hoffman}, and K.~{Saenko},
  ``Visda: A synthetic-to-real benchmark for visual domain adaptation,'' in
  \emph{Proceedings of the IEEE/CVF Conference on Computer Vision and Pattern
  Recognition Workshops}, 2018, pp. 2102--2105.

\bibitem{peng2019moment}
X.~Peng, Q.~Bai, X.~Xia, Z.~Huang, K.~Saenko, and B.~Wang, ``Moment matching
  for multi-source domain adaptation,'' in \emph{Proceedings of the IEEE/CVF
  International Conference on Computer Vision}, 2019, pp. 1406--1415.

\bibitem{saito2018maximum}
K.~Saito, K.~Watanabe, Y.~Ushiku, and T.~Harada, ``Maximum classifier
  discrepancy for unsupervised domain adaptation,'' in \emph{Proceedings of the
  IEEE conference on computer vision and pattern recognition}, 2018, pp.
  3723--3732.

\bibitem{zhang2019bridging}
Y.~Zhang, T.~Liu, M.~Long, and M.~Jordan, ``Bridging theory and algorithm for
  domain adaptation,'' in \emph{International Conference on Machine Learning},
  2019, pp. 7404--7413.

\bibitem{jiang2020implicit}
X.~Jiang, Q.~Lao, S.~Matwin, and M.~Havaei, ``Implicit class-conditioned domain
  alignment for unsupervised domain adaptation,'' in \emph{International
  Conference on Machine Learning}, 2020, pp. 4816--4827.

\bibitem{na2021fixbi}
J.~Na, H.~Jung, H.~J. Chang, and W.~Hwang, ``Fixbi: Bridging domain spaces for
  unsupervised domain adaptation,'' in \emph{Proceedings of the IEEE/CVF
  Conference on Computer Vision and Pattern Recognition}, 2021, pp. 1094--1103.

\bibitem{du2021cross}
Z.~Du, J.~Li, H.~Su, L.~Zhu, and K.~Lu, ``Cross-domain gradient discrepancy
  minimization for unsupervised domain adaptation,'' in \emph{Proceedings of
  the IEEE/CVF Conference on Computer Vision and Pattern Recognition}, 2021,
  pp. 3937--3946.

\bibitem{hu2020unsupervised}
L.~Hu, M.~Kan, S.~Shan, and X.~Chen, ``Unsupervised domain adaptation with
  hierarchical gradient synchronization,'' in \emph{Proceedings of the IEEE/CVF
  Conference on Computer Vision and Pattern Recognition}, 2020, pp. 4043--4052.

\bibitem{cui2020gradually}
S.~Cui, S.~Wang, J.~Zhuo, C.~Su, Q.~Huang, and Q.~Tian, ``Gradually vanishing
  bridge for adversarial domain adaptation,'' in \emph{Proceedings of the
  IEEE/CVF conference on computer vision and pattern recognition}, 2020, pp.
  12\,455--12\,464.

\bibitem{gu2020spherical}
X.~Gu, J.~Sun, and Z.~Xu, ``Spherical space domain adaptation with robust
  pseudo-label loss,'' in \emph{Proceedings of the IEEE/CVF Conference on
  Computer Vision and Pattern Recognition}, 2020, pp. 9101--9110.

\bibitem{lee2019sliced}
C.-Y. Lee, T.~Batra, M.~H. Baig, and D.~Ulbricht, ``Sliced wasserstein
  discrepancy for unsupervised domain adaptation,'' in \emph{Proceedings of the
  IEEE/CVF Conference on Computer Vision and Pattern Recognition}, 2019, pp.
  10\,285--10\,295.

\bibitem{miyato2018virtual}
T.~Miyato, S.-i. Maeda, M.~Koyama, and S.~Ishii, ``Virtual adversarial
  training: a regularization method for supervised and semi-supervised
  learning,'' \emph{IEEE transactions on pattern analysis and machine
  intelligence}, vol.~41, no.~8, pp. 1979--1993, 2018.

\bibitem{berthelot2019mixmatch}
D.~Berthelot, N.~Carlini, I.~Goodfellow, N.~Papernot, A.~Oliver, and C.~A.
  Raffel, ``Mixmatch: A holistic approach to semi-supervised learning,''
  \emph{Advances in Neural Information Processing Systems}, vol.~32, 2019.

\bibitem{sohn2020fixmatch}
K.~Sohn, D.~Berthelot, N.~Carlini, Z.~Zhang, H.~Zhang, C.~A. Raffel, E.~D.
  Cubuk, A.~Kurakin, and C.-L. Li, ``Fixmatch: Simplifying semi-supervised
  learning with consistency and confidence,'' \emph{Advances in Neural
  Information Processing Systems}, vol.~33, pp. 596--608, 2020.

\bibitem{Zou_2018_ECCV}
Y.~Zou, Z.~Yu, B.~Vijaya~Kumar, and J.~Wang, ``Unsupervised domain adaptation
  for semantic segmentation via class-balanced self-training,'' in
  \emph{European conference on computer vision}, 2018, pp. 289--305.

\bibitem{shu2018dirt}
R.~Shu, H.~Bui, H.~Narui, and S.~Ermon, ``A dirt-t approach to unsupervised
  domain adaptation,'' in \emph{International Conference on Learning
  Representations}, 2018.

\bibitem{zou2019confidence}
Y.~Zou, Z.~Yu, X.~Liu, B.~Kumar, and J.~Wang, ``Confidence regularized
  self-training,'' in \emph{Proceedings of the IEEE/CVF International
  Conference on Computer Vision}, 2019, pp. 5982--5991.

\bibitem{xu2019larger}
R.~Xu, G.~Li, J.~Yang, and L.~Lin, ``Larger norm more transferable: An adaptive
  feature norm approach for unsupervised domain adaptation,'' in
  \emph{Proceedings of the IEEE/CVF International Conference on Computer
  Vision}, 2019, pp. 1426--1435.

\bibitem{lu2020stochastic}
Z.~Lu, Y.~Yang, X.~Zhu, C.~Liu, Y.-Z. Song, and T.~Xiang, ``Stochastic
  classifiers for unsupervised domain adaptation,'' in \emph{Proceedings of the
  IEEE/CVF Conference on Computer Vision and Pattern Recognition}, 2020, pp.
  9111--9120.

\bibitem{prabhu2021sentry}
V.~Prabhu, S.~Khare, D.~Kartik, and J.~Hoffman, ``Sentry: Selective entropy
  optimization via committee consistency for unsupervised domain adaptation,''
  in \emph{Proceedings of the IEEE/CVF International Conference on Computer
  Vision}, 2021, pp. 8558--8567.

\end{thebibliography}
}

\end{document}